\documentclass[11pt]{article}
\usepackage[margin=1in]{geometry}
\usepackage{booktabs}
\usepackage{amsmath}
\usepackage{amssymb}
\usepackage{graphicx}
\usepackage{url}
\usepackage{natbib}
\usepackage{array}
\usepackage{longtable}
\usepackage{fvextra}
\usepackage[hidelinks]{hyperref}

\newcommand{\BlindCfPairs}{30}
\newcommand{\BlindInvPairs}{15}
\newcommand{\BlindN}{300}
\newcommand{\BlindSha}{090fedfed3660b34}

\newcommand{\BndCMiss}{80.0}

\newcommand{\BndDAnsTool}{17.5}
\newcommand{\BndDInv}{4.4}
\newcommand{\BndDMiss}{2.2}

\newcommand{\BndDRefEsc}{4.4}
\newcommand{\BndDToolConf}{20.0}
\newcommand{\BndHiAnsTool}{29.4}

\newcommand{\BndHiRefEsc}{12.6}
\newcommand{\BndHiToolConf}{34.3}
\newcommand{\BndLoAnsTool}{6.3}

\newcommand{\BndLoRefEsc}{-3.7}
\newcommand{\BndLoToolConf}{5.7}

\newcommand{\BndNMiss}{30}
\newcommand{\BndNPool}{128}

\newcommand{\BudgetSkewPct}{0.26}

\newcommand{\CiMethodCaption}{All paired confidence intervals for seed-averaged comparisons are 95\% percentile intervals from an item-clustered bootstrap: items (or counterfactual pairs, for pair metrics) are resampled with replacement and all three seeds of a resampled item move together, the statistic is the seed-averaged paired difference, $B=10{,}000$, seed 12345.}

\newcommand{\CleanLayers}{5}
\newcommand{\CleanNItems}{240}
\newcommand{\ConeBndN}{120}
\newcommand{\ConeBndNLayers}{3}

\newcommand{\ConeNSeeds}{3}
\newcommand{\ConeSpread}{8.7}

\newcommand{\ConemCzeroDiffSigned}{-1.0}

\newcommand{\ConfRouterAcc}{85.7}

\newcommand{\ConfRouterFrac}{50}

\newcommand{\ConfRouterTau}{0.978}
\newcommand{\CtrSeedDeltaOne}{22.9}
\newcommand{\CtrSeedDeltaThree}{0.8}
\newcommand{\CtrSeedDeltaTwo}{5.9}
\newcommand{\CzeroSpread}{1.7}
\newcommand{\EfourDiff}{4.9}

\newcommand{\EfourHi}{-1.7}
\newcommand{\EfourLo}{-8.1}
\newcommand{\EfourSeeds}{-7.7, -2.0, -5.0}
\newcommand{\EfourStable}{stable}
\newcommand{\EoneDiff}{11.7}

\newcommand{\EthreeDiffSigned}{-1.8}

\newcommand{\EthreeStable}{indeterminate}
\newcommand{\EtwoDiff}{16.6}

\newcommand{\FlipConMax}{0.50}
\newcommand{\FlipConMean}{0.29}
\newcommand{\FlipConMin}{0.18}
\newcommand{\FlipConPairs}{6}
\newcommand{\FlipJsonMax}{3.36}
\newcommand{\FlipJsonMean}{3.00}
\newcommand{\FlipJsonMin}{2.54}
\newcommand{\FlipJsonPairs}{10}
\newcommand{\FlipRelReduction}{90.3}
\newcommand{\GoldToolShare}{39.3}
\newcommand{\GoldTopAction}{CALL\_TOOL}
\newcommand{\JsonParseFail}{9}
\newcommand{\KfiveArms}{3}
\newcommand{\KfiveInBand}{0}
\newcommand{\KfiveSdHi}{13.5}
\newcommand{\KfiveSdLo}{1.0}
\newcommand{\LadderAlways}{88.0}
\newcommand{\LadderGap}{8.8}
\newcommand{\LadderNoThink}{70.7}
\newcommand{\LadderPointer}{74.6}
\newcommand{\LatPFiftyCut}{99.1}
\newcommand{\LatPNinetyFiveCut}{99.3}
\newcommand{\LayerMajorityHi}{60.0}
\newcommand{\LayerMajorityLo}{26.7}
\newcommand{\MajorityBaseline}{29.7}
\newcommand{\NUnitsCf}{30}

\newcommand{\NUnitsOvr}{300}
\newcommand{\NUnitsTool}{89}
\newcommand{\NUnitsToolRthree}{118}

\newcommand{\NonregDClean}{+1.37}
\newcommand{\NonregDFull}{+3.12}
\newcommand{\OneUnitCf}{3.3}
\newcommand{\OneUnitInv}{6.7}

\newcommand{\OpenDirCons}{5}
\newcommand{\OpenEasier}{3.9}
\newcommand{\OpenEasierHi}{5.2}
\newcommand{\OpenEasierLo}{2.6}
\newcommand{\OpenEfourDiff}{3.4}

\newcommand{\OpenEfourGrade}{Did not replicate}
\newcommand{\OpenEfourHi}{0.1}
\newcommand{\OpenEfourLo}{-6.8}
\newcommand{\OpenEfourStable}{indeterminate}
\newcommand{\OpenEnKoOff}{4.6}
\newcommand{\OpenEnKoTwo}{2.5}
\newcommand{\OpenEnKoTwoHi}{5.7}
\newcommand{\OpenEnKoTwoLo}{-0.7}
\newcommand{\OpenFmtOffDiff}{-4.8}
\newcommand{\OpenFmtOffHi}{0.4}
\newcommand{\OpenFmtOffLo}{-10.1}
\newcommand{\OpenFmtToolDiff}{2.6}
\newcommand{\OpenFmtToolHi}{10.5}
\newcommand{\OpenFmtToolLo}{-5.3}
\newcommand{\OpenKeptDropped}{22.4}
\newcommand{\OpenKeptDroppedHi}{33.3}
\newcommand{\OpenKeptDroppedLo}{11.7}
\newcommand{\OpenLabel}{1}
\newcommand{\OpenN}{248}
\newcommand{\OpenNDropped}{52}
\newcommand{\OpenNotRepl}{3}

\newcommand{\OpenOrigFmtToolDiff}{10.5}
\newcommand{\OpenOrigFmtToolHi}{18.4}
\newcommand{\OpenOrigFmtToolLo}{2.6}
\newcommand{\OpenPassClarify}{69}

\newcommand{\OpenPassMulti}{38}
\newcommand{\OpenRepl}{12}
\newcommand{\OpenReplNull}{6}
\newcommand{\OpenRows}{27}
\newcommand{\OpenThinkDiff}{21.8}
\newcommand{\OpenThinkHi}{27.8}
\newcommand{\OpenThinkLo}{16.1}
\newcommand{\OpenVzeroDir}{4}
\newcommand{\OpenVzeroNot}{3}
\newcommand{\OpenVzeroNull}{5}
\newcommand{\OpenVzeroRepl}{15}
\newcommand{\PathTokJsonMed}{542}
\newcommand{\PathTokJsonPNinetyFive}{1750}

\newcommand{\PathTokOneMed}{1}
\newcommand{\PfourAjson}{0.0}
\newcommand{\PfourAoneTok}{91.0}
\newcommand{\PfourBjson}{87.0}
\newcommand{\PfourBnone}{291}
\newcommand{\PfourBoneTok}{1.0}
\newcommand{\PfourLongTok}{964}
\newcommand{\PfourShortTok}{280}
\newcommand{\PfourTokRatio}{3.4}
\newcommand{\PriorCtrlTool}{11.1}
\newcommand{\PriorToolRatio}{3.4}
\newcommand{\PriorTreatTool}{37.8}
\newcommand{\REnAgree}{93.9}
\newcommand{\REnBndMagOk}{4}
\newcommand{\REnBndN}{4}
\newcommand{\REnBoth}{86.1}
\newcommand{\REnCiTol}{-8}
\newcommand{\REnDocEnFr}{1.28}
\newcommand{\REnDocEnFrHi}{0.35}
\newcommand{\REnDocEnFrLo}{-3.15}

\newcommand{\REnDocKoFr}{0.82}
\newcommand{\REnDocKoFrHi}{0.93}
\newcommand{\REnDocKoFrLo}{-2.68}

\newcommand{\REnEnOn}{88.34}

\newcommand{\REnFloorTol}{-5}
\newcommand{\REnFrEnDocOff}{1.51}
\newcommand{\REnFrEnDocOffHi}{3.03}
\newcommand{\REnFrEnDocOffLo}{0.12}

\newcommand{\REnGateFour}{passed}
\newcommand{\REnGateThree}{failed}
\newcommand{\REnGateTwo}{passed}

\newcommand{\REnKoOn}{89.16}

\newcommand{\REnN}{286}
\newcommand{\REnRenderTotal}{300}
\newcommand{\REnSelOff}{2.2}
\newcommand{\REnSelOn}{1.5}
\newcommand{\REnTCEnOff}{2.91}
\newcommand{\REnTCEnOffHi}{5.36}
\newcommand{\REnTCEnOffLo}{0.58}
\newcommand{\REnTCEnOffSeeds}{+5.2, -0.3, +3.9}

\newcommand{\REnTCEnOffStable}{indeterminate}
\newcommand{\REnTCEnOn}{0.00}

\newcommand{\REnTCKoOff}{3.03}
\newcommand{\REnTCKoOffHi}{6.18}
\newcommand{\REnTCKoOffLo}{0.00}

\newcommand{\REnTCKoOffStable}{indeterminate}
\newcommand{\REnTCKoOn}{0.93}

\newcommand{\REnXone}{0.35}

\newcommand{\REnXseven}{0.47}

\newcommand{\REnXsix}{0.00}

\newcommand{\REnXthree}{0.47}

\newcommand{\REnXthreeb}{1.28}

\newcommand{\REnXtwo}{0.82}
\newcommand{\REnXtwoHi}{0.93}
\newcommand{\REnXtwoLo}{-2.68}

\newcommand{\RaceCon}{70.7}

\newcommand{\RaceJson}{85.3}

\newcommand{\RacePtr}{74.6}

\newcommand{\RacePtrMax}{80.3}
\newcommand{\RacePtrMin}{67.0}
\newcommand{\RacePtrNSeeds}{3}
\newcommand{\RacePtrSd}{6.8}
\newcommand{\RecallDefAPp}{15.7}
\newcommand{\RecallDefBPp}{9.9}
\newcommand{\RecallDefCPp}{2.8}
\newcommand{\RecallDefSpread}{5.6}
\newcommand{\RecordsCtrl}{6{,}666}
\newcommand{\RecordsTreat}{6{,}588}
\newcommand{\RestartsCon}{4}
\newcommand{\RestartsJson}{5}
\newcommand{\RfourCThink}{88.7}

\newcommand{\RfourGapOff}{3.1}
\newcommand{\RfourNSeeds}{3}

\newcommand{\RfourVerdict}{substitution}
\newcommand{\RthreeCAcc}{74.4}

\newcommand{\RthreeCTool}{60.2}
\newcommand{\RthreeDAcc}{4.8}
\newcommand{\RthreeDCf}{8.9}
\newcommand{\RthreeDTool}{9.9}
\newcommand{\RthreeLoAcc}{2.00}
\newcommand{\RthreeLoCf}{-1.11}
\newcommand{\RthreeLoTool}{4.24}
\newcommand{\RthreeNSeeds}{3}

\newcommand{\RthreeRelTool}{24.8}

\newcommand{\RthreeSeedsAcc}{+9.0, +2.0, +3.3}
\newcommand{\RthreeSeedsCf}{+20.0, +0.0, +6.7}
\newcommand{\RthreeSeedsTool}{+22.9, +5.9, +0.8}
\newcommand{\RthreeTAcc}{79.2}

\newcommand{\RthreeTTool}{70.1}

\newcommand{\RzeroConClean}{96.75}
\newcommand{\RzeroConFull}{94.14}
\newcommand{\RzeroJsonClean}{96.86}
\newcommand{\RzeroJsonFull}{93.46}
\newcommand{\RzeroNClean}{2414}
\newcommand{\RzeroNFull}{2800}

\newcommand{\SensCleanFormatOffAbs}{0.8}

\newcommand{\SensCleanReason}{20.8}

\newcommand{\SensCleanTminusC}{5.4}
\newcommand{\SensFullFormatOff}{0.0}

\newcommand{\SensFullReason}{17.3}

\newcommand{\SensFullTminusC}{4.8}
\newcommand{\ShortcutDialCeil}{41.7}

\newcommand{\ShortcutDialReq}{73.3}

\newcommand{\ShortcutLayers}{7}
\newcommand{\ShortcutMissCeil}{51.7}
\newcommand{\ShortcutMissReq}{53.3}
\newcommand{\StepsCtrl}{417}
\newcommand{\StepsTreat}{412}

\newcommand{\TmConeDiff}{5.8}

\newcommand{\TmConeSeeds}{+9.7, +8.3, -0.7}
\newcommand{\TmConeStable}{indeterminate}
\newcommand{\TmCzeroDiff}{4.8}

\newcommand{\TokCodesVisible}{62}
\newcommand{\UaucConf}{0.743}
\newcommand{\UaucDisagree}{0.922}

\newcommand{\XFormatCfDiff}{-10.0}
\newcommand{\XFormatCfHi}{-0.0}
\newcommand{\XFormatCfLo}{-20.0}
\newcommand{\XFormatOffDiff}{0.0}
\newcommand{\XFormatOffHi}{4.0}
\newcommand{\XFormatOffLo}{-4.3}
\newcommand{\XFormatOnDiff}{2.7}
\newcommand{\XFormatOnHi}{6.0}
\newcommand{\XFormatOnLo}{-0.7}
\newcommand{\XFormatToolDiff}{12.4}
\newcommand{\XFormatToolHi}{20.2}
\newcommand{\XFormatToolLo}{4.5}

\newcommand{\XOffJsonTool}{50.6}
\newcommand{\XOffTok}{70.7}

\newcommand{\XOffTokTool}{62.9}

\newcommand{\XOnTok}{88.0}

\newcommand{\XOnTokTool}{93.3}
\newcommand{\XThinkBndJsonDiff}{16.4}
\newcommand{\XThinkBndJsonHi}{24.2}
\newcommand{\XThinkBndJsonLo}{8.6}
\newcommand{\XThinkBndTokDiff}{14.8}
\newcommand{\XThinkBndTokHi}{23.4}
\newcommand{\XThinkBndTokLo}{7.0}
\newcommand{\XThinkCfDiff}{43.3}
\newcommand{\XThinkCfHi}{63.3}
\newcommand{\XThinkCfLo}{23.3}
\newcommand{\XThinkDiff}{17.3}
\newcommand{\XThinkHi}{23.0}
\newcommand{\XThinkInvJsonDiff}{6.7}

\newcommand{\XThinkInvTokDiff}{0.0}
\newcommand{\XThinkInvTokHi}{20.0}
\newcommand{\XThinkInvTokLo}{-20.0}
\newcommand{\XThinkJsonDiff}{14.7}
\newcommand{\XThinkJsonHi}{19.7}
\newcommand{\XThinkJsonLo}{10.0}
\newcommand{\XThinkLo}{12.3}

\newcommand{\AbsJsonPFifty}{8.89}
\newcommand{\AbsJsonPNinetyFive}{27.43}
\newcommand{\AbsJsonTokMed}{538}
\newcommand{\AbsLatN}{40}
\newcommand{\AbsLatReps}{4}
\newcommand{\AbsOnePFifty}{85}
\newcommand{\AbsOnePNinetyFive}{168}
\newcommand{\CctConeMacroF}{72.9}
\newcommand{\CctConePrec}{91.2}
\newcommand{\CctConePredRate}{18.3}
\newcommand{\CctConeRec}{56.2}
\newcommand{\CctCzeroMacroF}{72.8}
\newcommand{\CctCzeroPrec}{91.9}
\newcommand{\CctCzeroPredRate}{19.1}
\newcommand{\CctCzeroRec}{59.2}
\newcommand{\CctGoldN}{89}
\newcommand{\CctGoldRate}{29.7}
\newcommand{\CctNItems}{300}
\newcommand{\CctNSeeds}{3}

\newcommand{\CctTConeMacroFSeeds}{${+}6.8$, ${+}4.6$, ${-}1.8$}

\newcommand{\CctTConePrecSeeds}{${-}7.0$, ${-}1.0$, ${+}4.2$}

\newcommand{\CctTConeRecSeeds}{${+}38.2$, ${+}19.1$, ${-}1.1$}

\newcommand{\CctTCzeroMacroFSeeds}{${+}6.6$, ${+}1.6$, ${+}1.8$}

\newcommand{\CctTCzeroPrecSeeds}{${+}0.6$, ${-}9.0$, ${+}2.5$}

\newcommand{\CctTCzeroRecSeeds}{${+}30.3$, ${+}9.0$, ${+}7.9$}
\newcommand{\CctTMacroF}{76.1}
\newcommand{\CctTPrec}{89.9}
\newcommand{\CctTPredRate}{24.8}
\newcommand{\CctTRec}{74.9}
\newcommand{\MsGatePos}{90}

\newcommand{\MsN}{300}
\newcommand{\MsPrimaryDiff}{${-}0.3$}
\newcommand{\MsPrimaryHi}{0.0}
\newcommand{\MsPrimaryLo}{-1.0}
\newcommand{\MsRealJson}{86.0}
\newcommand{\MsRealTok}{85.7}

\newcommand{\MsWrongJson}{50.0}
\newcommand{\MsWrongTok}{10.3}
\newcommand{\OpenKoSealN}{248}
\newcommand{\OpenKoShaPrefix}{48763eb3b50febaf}

\title{When Should Enterprise Policy Decisions Be\\
Learned, and When Should They Be Reasoned?}

\author{Hyojung Han$^{1,\dagger}$ \qquad Ryangkyung Kang$^{1}$ \qquad Youngeun Kim$^{2}$
\qquad Seung-Hun Jeon$^{1}$\\[4pt]
\small $^{1}$ThakiCloud \qquad $^{2}$Korea University\\[2pt]
\small $^{\dagger}$Corresponding author: \texttt{hyojung.han@thakicloud.com}}

\date{2026-10-02}

\begin{document}
\maketitle

\begin{abstract}
An enterprise agent that answers under a written policy has to choose an action
before it says anything: answer from the policy, ask for a missing fact, ask the
user to confirm a side effect, call one tool, call several, escalate to a human,
or refuse. We ask what that choice actually needs. Holding one 27B Korean
enterprise model fixed, we first study two factors: whether the decision is
produced by generating a JSON object or by reading a single constrained token,
and whether the model is allowed to reason before deciding. On one 27B Korean
enterprise model, we did not detect an accuracy difference between the two
operational paths on the in-distribution split (\RzeroJsonFull\,\% against
\RzeroConFull\,\% on \RzeroNFull\ items), while the two paths differ by two orders of
magnitude in latency and by roughly tenfold in the restart-to-restart
action-flip rate (\FlipJsonMean\,\% of actions flip against \FlipConMean\,\%). We are
careful about what that second comparison licenses: the two paths differ in
reasoning behaviour as well as in output format, emitting a median of
\PathTokJsonMed\ and \PathTokOneMed\ tokens respectively, so we report it as an
operational comparison and not as the cost of emitting the action. That result does not survive contact with policies the model has
not seen. On a sealed benchmark of \BlindN\ items whose labels were computed by a
hidden executable policy rather than written by anyone, a two-by-two separates
the two factors: in the single-token column reasoning is
worth ${+}\XThinkDiff\,$pp $[{+}\XThinkLo, {+}\XThinkHi]$ on exact action
(${+}\XThinkJsonDiff\,$pp in the generated-JSON column). In the reasoning-off row
the estimated exact-action difference between the two interfaces was
$\XFormatOffDiff\,$pp $[\XFormatOffLo, {+}\XFormatOffHi]$; this is a failure to
detect an advantage for either interface, not an equivalence result. With reasoning
on, the two arms do not share a reasoning prompt, so the format contrast there
(${+}\XFormatOnDiff\,$pp $[\XFormatOnLo, {+}\XFormatOnHi]$) is unresolved, and the
one secondary endpoint on which format mattered, a single-token advantage on
CALL\_TOOL-only recall, did not reproduce under an independent re-rendering of the
same scenarios. We then ask whether
the reasoning requirement is really a supervision gap. Against a control trained
on the same token budget of random in-domain data, boundary-targeted supervision
moves the two boundaries it explicitly covers by ${+}\BndDAnsTool$ and
${+}\BndDToolConf\,$pp, raises any-tool recall by ${+}\RthreeDTool\,$pp, and
does not detectably move an untouched held-out boundary
($[{\BndLoRefEsc}, {+}\BndHiRefEsc]\,$pp). Completing that design as a
training ${\times}$ reasoning factorial over both arms and \RfourNSeeds\ seeds, on
the sealed rendering for this model the interaction is negative and consistent
with partial substitution: reasoning is worth
${+}\EoneDiff\,$pp on the targeted arm against ${+}\EtwoDiff\,$pp on the random
one, an interaction of ${-}\EfourDiff\,$pp $[\EfourLo, \EfourHi]$ with the same
sign in every seed pair. Targeted data earns its keep where reasoning cannot be
afforded, not on top of it; the two are partial substitutes rather than
independent sources of accuracy. In the cleanly controlled reasoning-off
condition we did not detect a benefit from emitting the action as a generated
object, and a post-hoc attempt to isolate that interface with reasoning on failed because
the output contract changed the reasoning trace itself by a factor of
\PfourTokRatio. All of this supervision was Korean; applied with no
English training to independently rendered English versions of the same executable
scenarios, English is \REnXtwo\,pp lower $[\REnXtwoLo, {+}\REnXtwoHi]$ with
reasoning on, the same order as re-rendering the scenarios again in Korean. The
reasoning-off gain from targeted supervision has similar point estimates in the two
languages (${+}\REnTCKoOff$ in Korean, ${+}\REnTCEnOff\,$pp in English), but under
item-clustered intervals it is not stable in either, so we do not claim that it
carries over. The original rendering stays private as the confirmatory benchmark. We
re-rendered the same executable scenarios with open models after these results were
known and repeated every analysis: the reasoning effect and the English-transfer
verdict reproduced; the training-by-reasoning interaction, stable on the sealed set,
kept its sign but did not replicate under item-clustered intervals. We release the
model and that open re-rendering with its gold labels and proofs; the training corpus
remains internal.
\end{abstract}

\section{Introduction}

A deployed enterprise assistant is usually judged on what it says. Before it says
anything, though, it has already made the decision that matters: whether this
request can be answered from the policy document at all, whether a fact is
missing, whether a side effect needs confirmation, whether a tool has to run,
whether a human has to be involved, or whether the request must be refused. If
that choice is wrong, fluent text makes it worse rather than better.

Current practice answers that choice by generating it. The model emits a JSON
object naming an action and its arguments, and a downstream system executes it.
Generation is how we get structure, so generation is what we pay for, in latency
and in run-to-run variability. It is worth asking what the payment buys.

This paper separates two things that are usually bundled together. The first is
the \emph{output interface}: does the decision come out as a generated object, or
as a single token read from a constrained vocabulary? The second is
\emph{test-time reasoning}: is the model allowed to think before it decides? Both
are commonly removed together when a team optimizes an agent for cost, and when
quality drops the loss is attributed to whichever one the team happened to be
thinking about.

We find that the two behave completely differently, and that which one matters
depends on whether the policy is one the model has seen.

\paragraph{Contributions.}
\begin{enumerate}
\item On in-distribution enterprise policies, we did not detect an accuracy
difference between the single-token path and the deployed reasoning-plus-JSON path,
which the single-token path achieves at a fraction of the cost and with roughly
\FlipRelReduction\,\% less restart-to-restart instability. We report this
as a comparison of two operational paths, and we quantify the confound that stops
it from being a clean estimate of what emitting the action costs
(Section~\ref{sec:indist}).
\item We build a sealed decision benchmark whose gold labels are computed, not
written. A hidden executable policy maps latent factors to exactly one correct
action and emits a proof; natural-language renderings are produced separately and
are never allowed to see the label. No human and no language model ever supplied
a gold answer (Section~\ref{sec:bench}).
\item On that benchmark a two-by-two shows that reasoning drives the
exact-action generalisation gap, and that with reasoning off we detect no effect
of the output format on exact-action or pooled-boundary accuracy --- a failure to
detect, not an equivalence result --- reversing the ordering an in-distribution
evaluation would have suggested. With reasoning on, the format contrast is
confounded by the reasoning prompt and remains unresolved
(Section~\ref{sec:twobytwo}).
\item In a pre-registered comparison against a token-budget-matched random
control, boundary-targeted supervision improves decisions on the
boundaries it explicitly covers, and we detect no transfer to an untouched boundary
(Section~\ref{sec:r3}).
\item Completing the training ${\times}$ reasoning factorial on one harness, we
estimate the interaction directly and find it negative and stable on the sealed
rendering: supervision and test-time reasoning \textbf{substitute}
(Section~\ref{sec:ladder}). On the open re-rendering it keeps its sign but is not
stable under item-clustered intervals (Section~\ref{sec:openrerender}), so this
finding rests on the sealed set alone.
\item With no English supervision, we re-render the same latent scenarios in English
and find no English degradation at the resolution of a pre-registered audit; the
point estimates of the targeted-supervision gain without reasoning, and of its
disappearance with reasoning, are similar in English, but the reasoning-off gain is
not stable in either language (Section~\ref{sec:english}).
\end{enumerate}

\section{The decision task and three interfaces}
\label{sec:task}

The model sees a written policy, a list of callable tools with their signatures,
whatever the system has already established about the current session, and a user
turn. It must choose exactly one of seven actions: \textsc{answer},
\textsc{ask\_clarification}, \textsc{ask\_confirmation}, \textsc{call\_tool},
\textsc{call\_multiple\_tools}, \textsc{escalate}, \textsc{refuse}.

The base model is a 27B Korean enterprise assistant fine-tuned for this setting.
Sections~\ref{sec:indist}--\ref{sec:twobytwo} hold that checkpoint fixed and
change only how the decision leaves the model and whether reasoning is enabled;
from Section~\ref{sec:r3} onward the comparisons involve separately
trained checkpoints, and each section says which.

\begin{description}
\item[JSON generation.] The model autoregressively emits an object naming the
action. This is the deployed baseline.
\item[Constrained single token.] The action vocabulary is mapped to seven single
tokens and the decision is the argmax over that restricted set at one position.
No \emph{action} tokens are emitted; whether reasoning tokens precede the
decision is a separate switch that we vary deliberately.
\item[LoRA pointer head.] A rank-16 adapter plus a small head reads a hidden
state and scores the seven actions directly.
\end{description}

All three receive the same task content. Their output contracts differ, and under
reasoning-on those contracts induce different reasoning prompts and traces. The first two share all weights; the third adds
parameters.

Two different things are called ``generation'' in this literature and we keep
them apart throughout. \textbf{Action-output generation} is emitting the action
object itself, token by token. \textbf{Reasoning generation} is emitting tokens
before the decision that are never shown to the downstream system. Our claim
concerns the first, and it is narrower than ``unnecessary'': in the one condition
we can control cleanly --- reasoning off --- action-output generation brings no
advantage we can measure. The second is the one that turns out to matter, and
conflating them is how we nearly drew the wrong conclusion twice
(Sections~\ref{sec:indist} and~\ref{sec:twobytwo}). They are also not as
separable as this paragraph makes them sound; Section~\ref{sec:posthoc} reports
the experiment in which we tried to separate them and could not.

\section{In distribution, we detect no accuracy difference between the cheap and deployed paths}
\label{sec:indist}

We first measure on the held-out split of the distribution the model was trained
for: \RzeroNFull\ labelled items, of which \RzeroNClean\ survive a strict
cleanliness filter. We serve the model several separate times and evaluate after
each restart, which lets us separate accuracy from stability.

What we are comparing here is two \emph{operational paths}, not two output
formats. The deployed path reasons and then emits a JSON action object. The cheap
path does neither: reasoning is off and the decision is one constrained token.
The gap in work done is not subtle --- the deployed path emits a median of
\PathTokJsonMed\ tokens per item (95th percentile \PathTokJsonPNinetyFive), and
the cheap path emits exactly \PathTokOneMed. Any difference we see below is the
difference between those two paths, and we are explicit throughout about what
that does and does not license.

We did not detect an accuracy difference between them. The deployed path reaches \RzeroJsonClean\,\% on
the clean split and \RzeroJsonFull\,\% on the full split; the cheap path reaches
\RzeroConClean\,\% and \RzeroConFull\,\%. That is the operationally interesting
result on its own: on policies the model was trained for, we did not detect an
accuracy benefit from several hundred tokens of reasoning and a generated action
object over a single constrained token. We ran no paired equivalence test, and the
two splits disagree in the sign of the small difference
(Appendix~\ref{app:caveats}), so this is a failure to detect a difference, not
evidence that the two paths are equally accurate.

Stability does separate them, and strongly. Across restart pairs of the same
checkpoint on the same hardware, the deployed path changes its decision on
\FlipJsonMean\,\% of items on average
(\FlipJsonMin--\FlipJsonMax\,\% over \FlipJsonPairs\ pairs from
\RestartsJson\ restarts), while the cheap path changes on \FlipConMean\,\%
(\FlipConMin--\FlipConMax\,\% over \FlipConPairs\ pairs from \RestartsCon\
restarts). The two ranges do not overlap. We report the distribution over pairs
rather than a pooled interval over items on purpose: pairs drawn from the same
restart are not independent, and pooling them would manufacture precision the
design does not support.

Latency separates them by about two orders of magnitude: moving from the deployed
path to the cheap one cuts end-to-end p50 by up to \LatPFiftyCut\,\% and p95 by up
to \LatPNinetyFiveCut\,\%. In absolute terms, at concurrency one on a cold prefix,
the deployed path takes \AbsJsonPFifty\,s at p50 and \AbsJsonPNinetyFive\,s at p95
and generates a median of \AbsJsonTokMed\ tokens on the latency sample, against
\AbsOnePFifty\ and \AbsOnePNinetyFive\,ms and one token for the single-token path
(medians over \AbsLatReps\ restarts of a \AbsLatN-item sample;
Table~\ref{tab:app-latency}). We state the percentages as upper bounds because they
are end-to-end measurements of our own serving setup, and a different deployment
would place the two paths differently.

\begin{table}[t]
\centering
\begin{tabular}{lrrrrrr}
\toprule
Operational path & Clean & Full & Restarts & Pairs & Flip mean & Flip range \\
\midrule
Reasoning ${+}$ JSON action & 96.86\% & 93.46\% & 5 & 10 & 3.00\% & 2.54--3.36\% \\
No reasoning, single token & 96.75\% & 94.14\% & 4 & 6 & 0.29\% & 0.18--0.50\% \\
\bottomrule
\end{tabular}

\caption{Two operational paths on the in-distribution split. \emph{Clean} is the
\RzeroNClean-item strict split, \emph{Full} the \RzeroNFull-item split.
\emph{Flip} is the fraction of items whose chosen action differs between two
restarts of the same checkpoint on the same hardware, as the mean and range over
restart pairs; pairs share restarts and are not independent, so no pooled
interval is given. The two paths differ in reasoning as well as in output format,
so these rows compare deployments, not output interfaces.}
\label{tab:indist}
\end{table}

\paragraph{What this does not show.}
It is tempting to read the stability and latency rows as the cost of generating
the action, and we initially did. They are not. Both differences are consistent
with being driven entirely by the \PathTokJsonMed\ reasoning tokens rather than
by the handful of action tokens, and this design cannot separate the two. An
experiment that fixed reasoning and varied only the output format would settle
it. Section~\ref{sec:twobytwo} gets half of that: with reasoning off it isolates
the output format cleanly, and there we did not detect a format contribution to
the exact-action difference (point estimate $\XFormatOffDiff\,$pp,
$[\XFormatOffLo, {+}\XFormatOffHi]$; not an equivalence result). With reasoning on it does not, because the two arms do
not share a reasoning prompt. Section~\ref{sec:posthoc} reports what happened
when we tried to close that half directly, by freezing one reasoning trace and
branching only the output interface: the attempt failed, and the reason it failed
is that the output instruction changes the trace. We also did not repeat the
restart measurement under matched reasoning, so the decomposition of the
stability gap remains open.

What this section does establish is narrower and still useful to a practitioner:
for policies the model has been trained on, we did not detect the expensive
deployed path buying accuracy, and the cheap path is markedly more reproducible
across restarts. It says nothing about a policy the model has never seen, which is the
situation every enterprise deployment is in on its second customer.

\section{A benchmark nobody labelled}
\label{sec:bench}

To ask the out-of-distribution question we needed a benchmark that is genuinely
unseen and whose labels we can defend. Two earlier attempts failed in ways worth
recording.

We first tried to label items drawn from a pool built on real statutes and internal
policy text by cross-vendor adjudication: two different commercial models labelled
each item independently. Agreement between them was poor on both pools we tried
($\kappa = 0.47$ on 1{,}193 items and $\kappa = 0.40$ on 853). Both runs also left
roughly 30\,\% of items abstained and neither completed its challenger stage, so we
treat these numbers as evidence that the approach was not converging rather than as
a clean reliability estimate.

The more telling signal was what the adjudicators chose. On the items meant to probe
the answer-versus-call-a-tool boundary, they chose \textsc{call\_tool} on 3 of 348,
against \textsc{answer} on 182. Layering tools and principles onto real legal text
does not put the intended boundary into the text. The boundary existed in our heads
and in our selection criteria, not in the item.

So we stopped writing items and started computing them. A hidden policy is
represented as a set of latent factors --- whether live state is needed, whether a
required fact is missing and whether the user can supply it, whether a side
effect requires confirmation, whether the request is prohibited and whether
escalation would make it permissible, whether human review is required, how many
tools are implicated, and whether the request is answerable from static policy
alone. A deterministic executor maps a factor assignment to exactly one action
under a fixed priority order, and emits a proof: the rule that fired and the
factors that made it fire. We enumerated the full factor space, kept the
assignments for which exactly one action is derivable, and sampled from those.

Natural-language renderings were then produced from the factor assignment by
models outside the evaluated family, and never saw the gold label. Machine gates
check that the rendering does not leak the answer, that no reference is left
unresolved, that counterfactual pairs differ in exactly one generating condition
and flip the gold action, and that invariance pairs differ only in surface dialect
and keep it. Two further gates check rendering fidelity with models outside both
the evaluated family and the renderer's vendor. In the \emph{recovery} gate, two
independent LLM extractors read only the rendered item and recover its decision
factors; a mismatch with the generating assignment on any decision factor rejects
the item. In the \emph{challenge} gate, a challenger model looks for an
alternative reading of the item under which a different action would be correct;
an upheld challenge rejects it. Items were selected using structural difficulty metadata only ---
exception depth, number of conditions, noise level, distractor count, tool count
--- and never using any model's performance on them.

\begin{figure}[p]
\centering
\includegraphics[width=\linewidth]{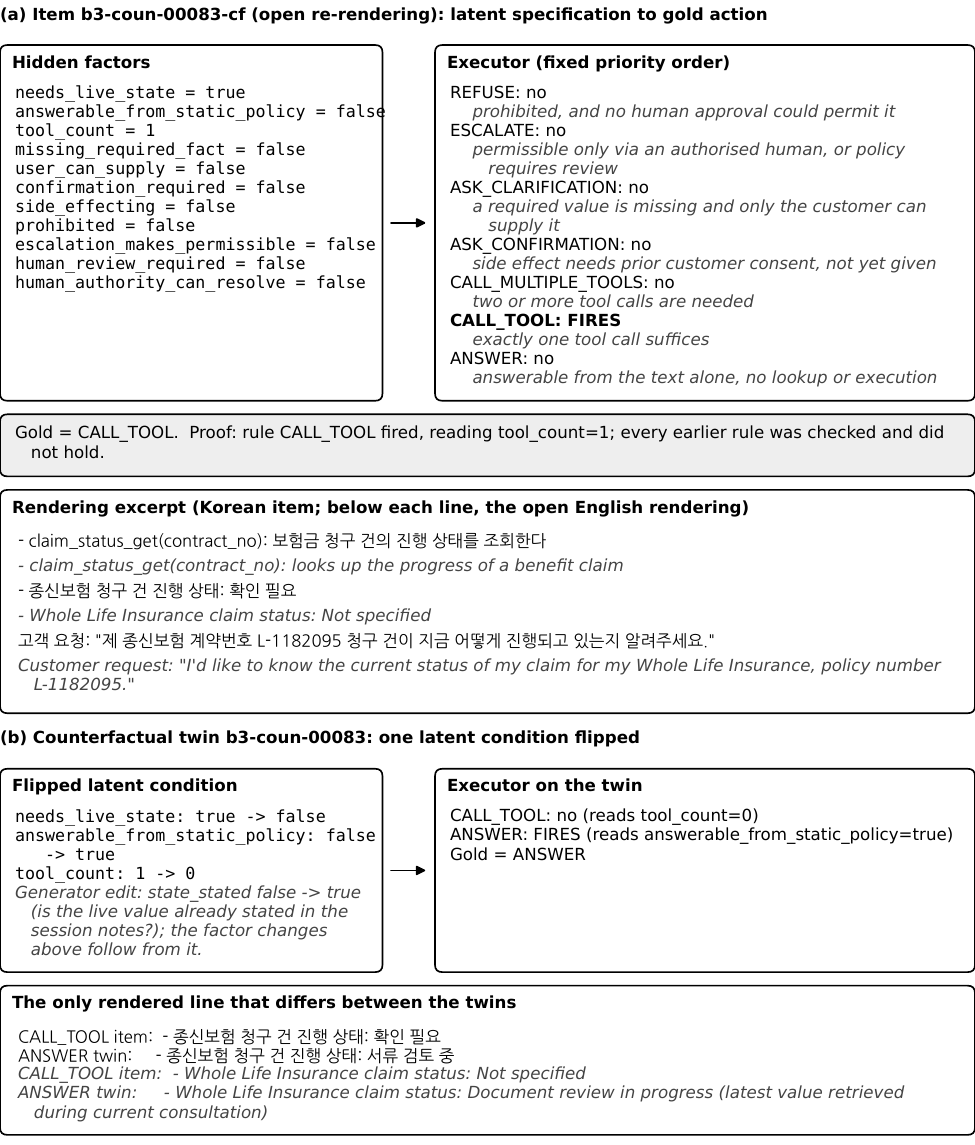}
\caption{One scenario of the public open re-rendering, end to end. (a) Hidden
factors of item \texttt{b3-coun-00083-cf}; the executor checks its rules in a fixed
priority order and the first that holds, \textsc{call\_tool}, is the gold action,
with a proof naming the rule and the factors it read. Below: three rendered Korean
lines, each followed by the independently rendered English line. (b) Its
counterfactual twin: one generating condition is flipped, the dependent factors
follow, the executor fires \textsc{answer}, and exactly one rendered line differs.
Long lines wrap with an indent.}
\label{fig:example}
\end{figure}

Figure~\ref{fig:example} walks one scenario of the public re-rendering through
this pipeline: the hidden factors, the executor's derivation, the gold action with
its proof, an excerpt of the rendered item, and the counterfactual twin in which one
generating condition is flipped and the gold action changes from
\textsc{call\_tool} to \textsc{answer}. The executor proves the action implied by
the latent specification; it does not by itself prove that the natural-language
rendering faithfully expresses that specification. Rendering fidelity is checked
separately by the recovery and challenge gates described above.

The result is \BlindN\ items across seven layers, including \BlindCfPairs\
counterfactual pairs and \BlindInvPairs\ invariance pairs, sealed by content hash
(\texttt{\BlindSha}\ldots) before any evaluated model was run against it. A
twelve-item mechanical checklist verifies all of the above and is re-run from the
sealed artifact rather than trusting the generator's own records.

\paragraph{Shortcut diagnostics, including the two that failed.}
A benchmark whose items are constructed invites the question of whether a model
can win without reading the policy. We ran lexical probes on the final \BlindN\
items: a majority-class baseline, a classifier given only the request, one given
only the policy, and one given both. Majority alone scores
\MajorityBaseline\,\% overall, and per layer it ranges from \LayerMajorityLo\,\%
to \LayerMajorityHi\,\% (the most frequent gold action is
\GoldTopAction), which is the floor every number in this paper should be read
against.

Five of the \ShortcutLayers\ layers come in under their own ceiling on every probe. Two do
not. On the dialect layer, a request-only probe reaches \ShortcutDialReq\,\% against a
ceiling of \ShortcutDialCeil\,\%, and policy-only and full-lexical probes also clear it; on
the missing-information layer a request-only probe reaches \ShortcutMissReq\,\% against a
ceiling of \ShortcutMissCeil\,\%. We report this rather than bury it, and it has a direct
consequence: all \BlindInvPairs\ invariance pairs live inside the dialect layer,
so the invariance endpoint is computed entirely within a slice where a lexical
shortcut exists. That is a second reason, alongside its resolution, not to rest
anything on it.

These probes ran after selection and were not fed back into it, which is a
weakness of the pipeline rather than of this particular set. The twelve-item
checklist in the seal covers leakage of the answer into the prompt; it does not
cover lexical shortcuts, and a future version should gate on both before sealing.

We stress what this does and does not buy. It buys a label we can defend without
appealing to anyone's judgement, and a counterfactual design in which the
single changed variable is known by construction rather than asserted. It does
not buy ecological validity: these are constructed policies, and a model could be
good at them while being bad at a real company's documents.

\paragraph{Public re-rendering.} The rendering described here, written by two
commercial models, remains private and is the confirmatory benchmark for every
primary analysis. For release we re-rendered the same latent scenarios and gold
labels with open-weight models after these results were known;
Section~\ref{sec:openrerender} describes that re-rendering and repeats the analysis
on it.

\section{On unseen boundaries, reasoning drives the generalisation gap}
\label{sec:twobytwo}

Our first run on the sealed set compared the three interfaces of
Section~\ref{sec:task} and produced an apparent reversal: generated JSON reached
\RaceJson\,\%, the pointer head \RacePtr\,\% (a \RacePtrNSeeds-seed mean, sd
\RacePtrSd\,pp across seeds), and the constrained single token \RaceCon\,\%. The
generated and single-token numbers are single runs; the pointer head is the only
multi-seed entry, and it is not part of Table~\ref{tab:twobytwo}, whose cells
are all single runs of one checkpoint. Read naively this says the generated object matters
after all, and matters by about fifteen points.

It says no such thing. The generated arm had test-time reasoning enabled and the
constrained arm did not. The comparison confounded the two factors we set out to
separate, and we would have published the wrong cause.

\begin{figure}[t]
\centering
\includegraphics[width=0.95\linewidth]{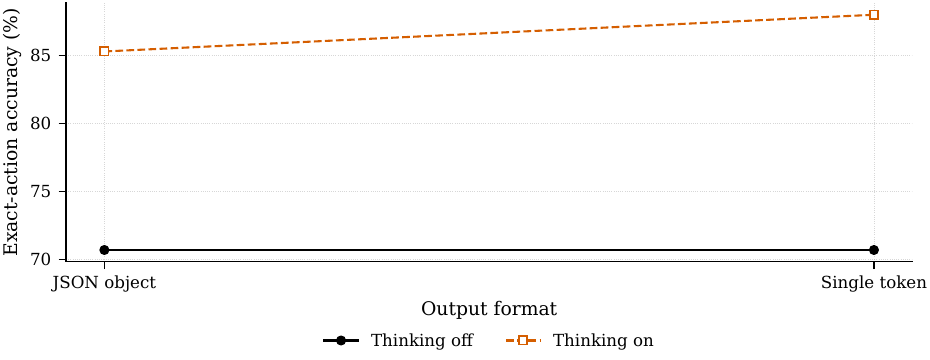}
\caption{Exact-action accuracy on the sealed benchmark. The vertical gap between
the two lines is test-time reasoning; the horizontal slope of either line is the
output format. The reasoning-off horizontal contrast isolates the output format
cleanly; the reasoning-on horizontal contrast retains the prompt confound
described in the text, because those two arms do not share a reasoning prompt.
Holding format fixed, reasoning is worth
${+}\XThinkDiff\,$pp $[{+}\XThinkLo, {+}\XThinkHi]$. Holding reasoning off,
changing the format is worth $\XFormatOffDiff\,$pp
$[\XFormatOffLo, {+}\XFormatOffHi]$; holding it on, ${+}\XFormatOnDiff\,$pp
$[\XFormatOnLo, {+}\XFormatOnHi]$. Format effects are written as single
token minus JSON. Intervals are paired item-resampling bootstraps over the
\NUnitsOvr\ items.}
\label{fig:twobytwo}
\end{figure}

We therefore ran the full two-by-two: reasoning on or off, crossed with generated
JSON or constrained single token, on the same sealed items
(Table~\ref{tab:twobytwo}, Figure~\ref{fig:twobytwo}).

\begin{table}[t]
\centering
\begin{tabular}{llrr}
\toprule
Metric & Reasoning & JSON object & Single token \\
\midrule
Exact action & off & 70.7\% & 70.7\% \\
             & on  & 85.3\% & 88.0\% \\
\addlinespace
Both-correct (CF pairs) & off & 46.7\% & 36.7\% \\
             & on  & 76.7\% & 80.0\% \\
\bottomrule
\end{tabular}

\caption{Reasoning crossed with output format on the sealed benchmark
($n=\BlindN$; \BlindCfPairs\ counterfactual pairs). Each cell is a single run of
the same checkpoint. Reasoning moves both metrics by a wide margin. With
reasoning off, we did not detect an output-format effect on exact action on this
sample (a failure to detect a difference, not an equivalence result); with reasoning on, the
two arms do not share a reasoning prompt, so the format contrast in that row is
unresolved. On the counterfactual endpoint the format effect is not identically
zero, and the one secondary endpoint where it is clearly non-zero favours the
single token on this rendering only (see text and
Section~\ref{sec:openrerender}).}
\label{tab:twobytwo}
\end{table}

\begin{figure}[t]
\centering
\includegraphics[width=0.95\linewidth]{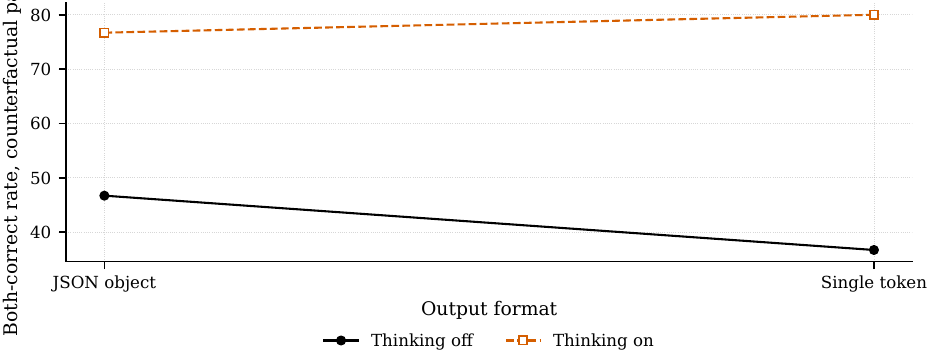}
\caption{The same two-by-two scored on counterfactual pairs, where an item counts
only if the model gets both members right. The effect of reasoning is larger here
(${+}\XThinkCfDiff\,$pp $[{+}\XThinkCfLo, {+}\XThinkCfHi]$) because getting both
members right requires responding to the one latent variable that changed rather
than to the surface of either item. Unlike exact action, the format effect here is
not identically zero ($\XFormatCfDiff\,$pp with reasoning off, interval
$[\XFormatCfLo, \XFormatCfHi]$), so we report it as inconclusive. This is a
different quantity from Figure~\ref{fig:twobytwo} and is therefore plotted on its
own axis.}
\label{fig:twobytwocf}
\end{figure}

The benchmark's scoring harness declares three primary endpoints --- pooled
boundary accuracy, both-correct on counterfactual pairs, and both-correct on
invariance pairs --- with overall exact action and \textsc{call\_tool}-only recall
($n=\NUnitsTool$ items whose gold action is exactly \textsc{call\_tool}) as
secondary. We report all five, because the one we would most have liked to leave
out is the one that disagrees with us.

Reasoning moves the two endpoints with enough resolution to move. Holding the
output format at a single token, it is worth ${+}\XThinkDiff\,$pp
$[{+}\XThinkLo, {+}\XThinkHi]$ on exact action, ${+}\XThinkBndTokDiff\,$pp
$[{+}\XThinkBndTokLo, {+}\XThinkBndTokHi]$ on pooled boundary accuracy, and
${+}\XThinkCfDiff\,$pp $[{+}\XThinkCfLo, {+}\XThinkCfHi]$ on counterfactual
pairs. Holding it at JSON instead, exact action gives ${+}\XThinkJsonDiff\,$pp
$[{+}\XThinkJsonLo, {+}\XThinkJsonHi]$ and pooled boundary accuracy
${+}\XThinkBndJsonDiff\,$pp $[{+}\XThinkBndJsonLo, {+}\XThinkBndJsonHi]$. We give
both columns rather than the larger one.

On invariance pairs we did not detect a reasoning effect. In the single-token column the effect
is exactly $\XThinkInvTokDiff\,$pp, with an interval of
$[\XThinkInvTokLo, {+}\XThinkInvTokHi]$; in the JSON column it is
${+}\XThinkInvJsonDiff\,$pp, also containing zero. With \BlindInvPairs\ pairs one
pair is worth \OneUnitInv\,pp, so this endpoint cannot resolve anything smaller
than a large effect, and we read it as uninformative rather than as evidence of
no effect. We flag it because it is a declared primary endpoint and omitting it
would make the separation look cleaner than it is.

For the output format, exact action with reasoning off gives
$\XFormatOffDiff\,$pp $[\XFormatOffLo, {+}\XFormatOffHi]$ and with reasoning on
${+}\XFormatOnDiff\,$pp $[\XFormatOnLo, {+}\XFormatOnHi]$. On pooled boundary
accuracy we did not detect a format difference in the reasoning-off row (its
interval contains zero). The reasoning-on interval also contains zero, but that row
is confounded by the unshared reasoning prompt (below), so we read it as unresolved,
not as a null. On counterfactual pairs with reasoning off
the single token is $\XFormatCfDiff\,$pp relative to generated JSON, with an
interval of $[\XFormatCfLo, \XFormatCfHi]$ that touches zero, and on invariance
pairs with reasoning on it is $-\XThinkInvJsonDiff\,$pp. (Throughout, a format
effect is written as single token minus JSON.) So the honest summary is that
we did not detect an output-format effect on exact action or pooled boundary
accuracy in the reasoning-off row --- the estimated exact-action difference is
$\XFormatOffDiff\,$pp $[\XFormatOffLo, {+}\XFormatOffHi]$, a failure to detect an
advantage for either interface rather than an equivalence result --- while on the
two pair-based endpoints the point estimates favour JSON by amounts smaller than
this benchmark can resolve. We draw no conclusion from those two.

The reasoning-on row is different in kind. There the two arms reason under
different instructions: the single-token arm reasons under the single-token
suffix, the JSON arm under the JSON contract (Section~\ref{sec:limitations}).
The post-hoc control of Section~\ref{sec:posthoc}, which tried to freeze one trace
and branch only the output interface, did not resolve this, because the output
instruction changed the trace itself. We therefore mark the reasoning-on format
effect as unresolved; only the reasoning-off format contrast is cleanly
identified.

The one place a format effect clearly excludes zero on this rendering is
\textsc{call\_tool}-only recall ($n=\NUnitsTool$) with reasoning off, and there it favours the single token: \XOffTokTool\,\% of
required tool calls against \XOffJsonTool\,\% for generated JSON, a difference of
${+}\XFormatToolDiff\,$pp $[{+}\XFormatToolLo, {+}\XFormatToolHi]$ on the
\NUnitsTool\ items whose gold action is exactly \textsc{call\_tool}. With reasoning
on, both formats sit above \XOnTokTool\,\% and we did not detect a difference. This effect did not
survive re-rendering: on the open rendering of the same scenarios it is
${+}\OpenFmtToolDiff\,$pp $[\OpenFmtToolLo, {+}\OpenFmtToolHi]$, while the original
rendering restricted to the same items keeps it (Section~\ref{sec:openrerender}). We
therefore report it as a property of this rendering, not as a format effect.

Two properties of the harness bound how far the absolute numbers in the reasoning
row can be pushed. The \textsc{on}-row JSON arm failed to produce a parseable
action on \JsonParseFail\ of \BlindN\ items, which we scored as errors; removing
them would close most of the format gap in that row. And in the reasoning-plus-
single-token arm, all seven action codes appeared in the top-20 logits for only
\TokCodesVisible\ of \BlindN\ items, so the remaining decisions were an argmax
over a visible subset. Both apply equally across the comparisons we draw from
this table, but they mean the cell values should not be carried over to settings
with a different harness.

So the in-distribution conclusion and the out-of-distribution conclusion are both
true and they are about different things. Only the reasoning-off row cleanly
identifies the output-format contrast. In the other regimes we observed no accuracy
advantage for generation, but those comparisons remain confounded and are not
interpreted as format effects. Test-time reasoning, by contrast, has no effect we could detect in one regime and
is decisive in the other.

A natural next move is to spend reasoning only where it is needed. A confidence
router that sends low-confidence items to the reasoning path, with the threshold
chosen on in-distribution data before the sealed set was opened
($\tau=\ConfRouterTau$), reaches \ConfRouterAcc\,\% while reasoning on
\ConfRouterFrac\,\% of items, against \LadderAlways\,\% for always reasoning and
\LadderNoThink\,\% for never reasoning. The saving is real but modest, and the
reason is visible in the failure mode: on unseen boundaries the non-reasoning
model is frequently confident and wrong, which is precisely the regime in which a
confidence signal cannot help.

\section{Is the reasoning requirement a supervision gap?}
\label{sec:r3}

The two-by-two shows that reasoning is needed. It does not show \emph{why}. Two
explanations are compatible with it. The boundaries may require serial
computation that a single forward pass cannot perform. Or the model may simply
never have been taught these boundaries, in which case targeted supervision
should close the gap and reasoning would be a workaround for missing training
data, not a computational necessity.

We pre-registered a test before scoring the sealed set. The treatment arm adds
boundary-targeted training records --- contrastive pairs that differ in one
decision-relevant factor across three boundaries: answer versus call a tool, call
a tool versus ask for confirmation, and missing information. The control arm adds
the \emph{same number of tokens} of randomly sampled in-domain data. Everything
else --- recipe, schedule, seeds --- is identical, so data is the only
independent variable. Each arm is trained with \RthreeNSeeds\ seeds. The
held-out boundary, refuse versus escalate, appears in neither arm's added data.

\begin{table}[t]
\centering
\begin{tabular}{lrrrrl}
\toprule
& $n$ & Random & Targeted & $\Delta$ & 95\% CI \\
\midrule
\multicolumn{6}{l}{\emph{Pre-registered primary endpoints}} \\
Exact action & 300 & 74.4\% & 79.2\% & ${+}4.8$ & lower $={+}2.00$\,pp \\
Both-correct (CF pairs) & 30 & 46.7\% & 55.6\% & ${+}8.9$ & lower $={-}1.11$\,pp \\
Any-tool recall & 118 & 60.2\% & 70.1\% & ${+}9.9$ & lower $={+}4.24$\,pp \\
\addlinespace
\multicolumn{6}{l}{\emph{Per boundary (reporting only)}} \\
ANSWER\,/\,CALL\_TOOL & 42 & 64.3\% & 81.7\% & ${+}17.5$ & $[+6.3, +29.4]$ \\
CALL\_TOOL\,/\,CONFIRM & 35 & 55.2\% & 75.2\% & ${+}20.0$ & $[+5.7, +34.3]$ \\
Missing information & 30 & 80.0\% & 82.2\% & ${+}2.2$ & $[-2.2, +7.8]$ \\
REFUSE\,/\,ESCALATE\ \textbf{(held out)} & 45 & 66.7\% & 71.1\% & ${+}4.4$ & $[-3.7, +12.6]$ \\
Pooled boundary groups (union) & 128 & 70.6\% & 78.9\% & ${+}8.3$ & $[+3.1, +13.8]$ \\
Both-correct (invariance pairs) & 15 & 88.9\% & 93.3\% & ${+}4.4$ & $[+0.0, +11.1]$ \\
\bottomrule
\end{tabular}

\caption{Boundary-targeted supervision against a token-budget-matched random
control, \RthreeNSeeds\ seeds per arm, evaluated on the sealed benchmark. Note
that $n$ differs by endpoint: \NUnitsOvr\ items overall, \NUnitsCf\
counterfactual pairs, \NUnitsToolRthree\ items where a tool call is required
(this endpoint's ledger counts \textsc{call\_multiple\_tools} as tool-requiring;
the two-by-two ledger of Section~\ref{sec:twobytwo} does not, and reports
\NUnitsTool). Primary
endpoints were fixed before the benchmark was scored; per-boundary rows are
reporting only. Every interval, including the pre-registered lower bounds, is the
item-clustered bootstrap used throughout the paper: items (or counterfactual
pairs) are resampled with replacement, the \RthreeNSeeds\ seeds of an item move
together, the statistic is the seed-averaged paired difference ($B=10{,}000$,
seed 12345, 95\% percentile). The registered gate reads only the lower end of that
interval, so the primary rows print the lower bound alone. The intervals carry
evaluation sampling uncertainty for these \RthreeNSeeds\ trained seed pairs, not
retraining uncertainty, which we report separately as per-seed deltas in the text.
The same exact-action contrast appears as $T-C0$ in Table~\ref{tab:c1} with the
same interval. The \textbf{held-out} row is a boundary that neither arm's
added data covers. \emph{Pooled boundary groups} is item accuracy over the union of
the four per-boundary rows above it (each item counted once, held-out boundary
included; $n=\BndNPool$), averaged over seeds. It is not the ``boundary layers''
column of Table~\ref{tab:c1}, which is defined differently.}
\label{tab:r3}
\end{table}

\begin{figure}[t]
\centering
\includegraphics[width=0.95\linewidth]{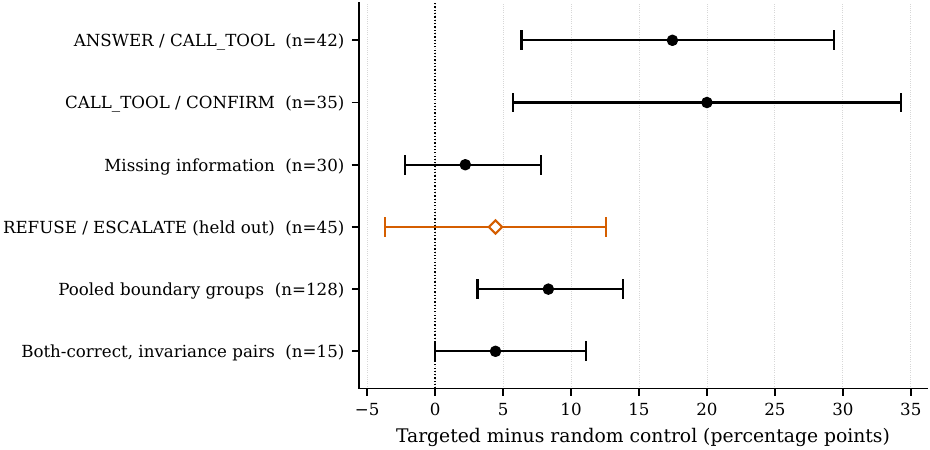}
\caption{Boundary-targeted supervision minus the token-budget-matched random
control, per boundary, with 95\% item-clustered bootstrap intervals (the
method of Table~\ref{tab:r3}) over the \RthreeNSeeds\ trained seed pairs. The two covered boundaries with headroom
clear zero; the held-out boundary (hollow diamond) does not. Missing information
barely moves: its control arm already sits at \BndCMiss\,\% on \BndNMiss\ items,
leaving little headroom. This is consistent with limited headroom rather than a
failure of the supervision, but the resolution here does not establish that on
its own, and we do not interpret the ${+}\BndDMiss\,$pp estimate.}
\label{fig:boundaries}
\end{figure}

The pre-registered decision rule asked for an absolute gain of at least 8\,pp or a
relative error reduction of at least 20\,\% on at least one primary endpoint,
with a paired interval excluding zero, together with non-regression in
distribution. One endpoint met it: any-tool recall ($n=\NUnitsToolRthree$ items
whose gold action is \textsc{call\_tool} or \textsc{call\_multiple\_tools}) rose from \RthreeCTool\,\% to
\RthreeTTool\,\%, a \RthreeRelTool\,\% relative error reduction with a lower
bound of ${+}\RthreeLoTool\,$pp. The seed pairs behind that single passing
endpoint are ${+}\CtrSeedDeltaOne$, ${+}\CtrSeedDeltaTwo$ and
${+}\CtrSeedDeltaThree\,$pp: all positive, but one pair carries most of the
effect and another is within a single item of nothing. The pre-registered bar
was written on the pooled estimate, and it was met; we print the spread so the
reader can see that the result rests on three runs, not on a stable margin. Overall accuracy rose by ${+}\RthreeDAcc\,$pp
with a lower bound of ${+}\RthreeLoAcc\,$pp --- consistent and positive across
all three seed pairs (\RthreeSeedsAcc\,pp), but below both pre-registered
effect-size thresholds. Counterfactual both-correct rose ${+}\RthreeDCf\,$pp with
a lower bound of $\RthreeLoCf\,$pp; across seeds the paired deltas were
\RthreeSeedsCf\,pp, so we report it as positive but inconclusive. In distribution
the registered non-regression check passed; the treatment arm was in fact better than the reference by
\NonregDClean\,pp on the clean split and \NonregDFull\,pp on the full split.

The per-boundary picture is the more informative one, and it is deliberately
reported as secondary. The two covered boundaries with headroom moved decisively:
answer versus call a tool by ${+}\BndDAnsTool\,$pp
$[{+}\BndLoAnsTool, {+}\BndHiAnsTool]$ and call a tool versus ask for
confirmation by ${+}\BndDToolConf\,$pp $[{+}\BndLoToolConf, {+}\BndHiToolConf]$.
The third covered boundary, missing information, moved by ${+}\BndDMiss\,$pp with
an interval containing zero; its control arm was already near the top of its
range, so we read this as a headroom limit rather than a failure, and we flag
that reading as an interpretation we did not pre-register.

The held-out boundary did not move in a way we can defend: ${+}\BndDRefEsc\,$pp
$[\BndLoRefEsc, {+}\BndHiRefEsc]$, an interval that contains zero. On the evidence
we have, boundary-targeted supervision improves decisions on the boundaries it is given
and we cannot claim it transfers to one it is not given.

\section{Learned or reasoned}
\label{sec:ladder}

\begin{figure}[t]
\centering
\includegraphics[width=0.95\linewidth]{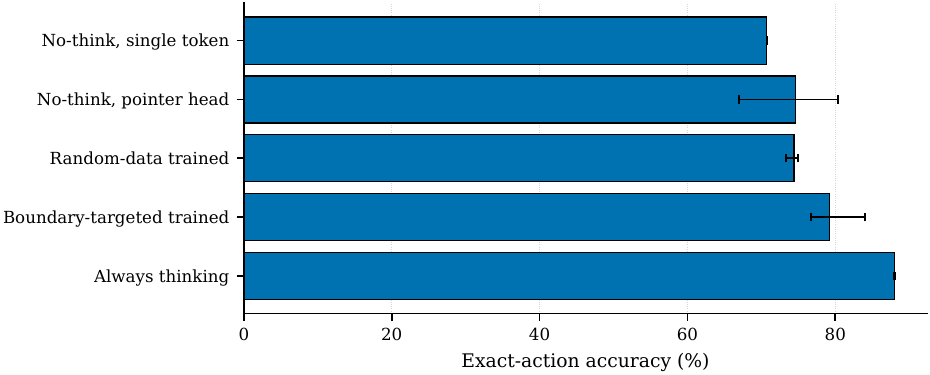}
\caption{Alternative configurations of data and test-time compute, scored on the
sealed benchmark. These are competing options, not a cumulative progression: the
pointer head and the random-data checkpoint are different artefacts, not one
built on the other. The first four spend a single forward pass; the last spends
test-time reasoning. The three trained rows --- the pointer head
(\RacePtrNSeeds\ seeds, \RacePtrMin--\RacePtrMax\,\%) and the two supervision arms
(\RthreeNSeeds\ seeds each) --- are seed means and carry a bar spanning their seed
minimum and maximum; the other two are single runs and carry no bar, because they
have no spread to show rather than a small one.
The trained bars overlap, which is the point of Section~\ref{sec:r3}'s
pre-registered reading: only any-tool recall cleared its bar. Two harnesses share
this axis: the first and last bars are the vLLM constrained-token probe of
Section~\ref{sec:twobytwo} on the base checkpoint, the three trained bars the
pointer-head harness of Section~\ref{sec:r3}.}
\label{fig:ladder}
\end{figure}

Putting the configurations side by side shows where the division of labour falls
(Figure~\ref{fig:ladder}, Table~\ref{tab:ladder}). Without reasoning, a
constrained decision reaches \LadderNoThink\,\%. A trained pointer head reaches
\LadderPointer\,\%. A token-budget-matched helping of random in-domain data
reaches \RthreeCAcc\,\%. Boundary-targeted pairs instead of that random data
reach \RthreeTAcc\,\%. The base model allowed to reason reaches \LadderAlways\,\%.

\begin{table}[t]
\centering
\begin{tabular}{lr}
\toprule
Configuration & Exact action \\
\midrule
No thinking, constrained single token & 70.7\% \\
No thinking, LoRA pointer head & 74.6\% \\
Trained on random in-domain data & 74.4\% \\
Trained on boundary-targeted data & 79.2\% \\
\addlinespace
Always thinking, constrained single token & 88.0\% \\
\bottomrule
\end{tabular}

\caption{The same configurations as numbers. Rows above the rule are single-pass;
the row below spends test-time reasoning. The pointer-head row is a
\RacePtrNSeeds-seed mean (sd \RacePtrSd\,pp) and the two supervision rows are
\RthreeNSeeds-seed means; the constrained-token and always-thinking rows are
single runs. The rows come from two harnesses: the first and last rows from the
vLLM constrained-token probe of Section~\ref{sec:twobytwo} on the base checkpoint,
the middle three from the pointer-head harness of Section~\ref{sec:r3}. They are
competing configurations measured by different decision mechanisms, not
differences of one another.}
\label{tab:ladder}
\end{table}

\begin{figure}[t]
\centering
\includegraphics[width=0.95\linewidth]{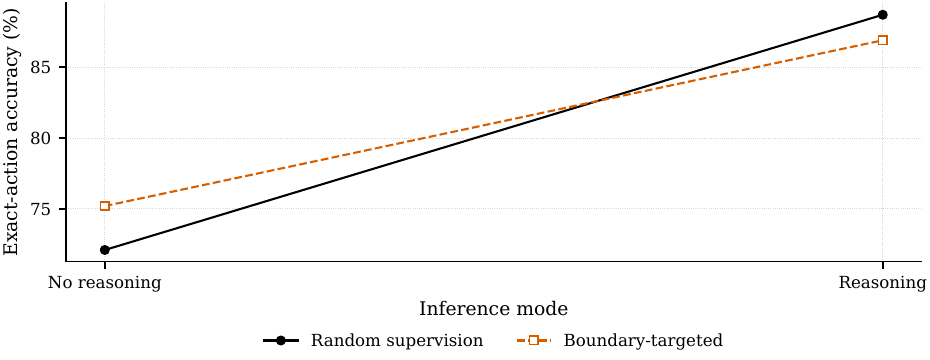}
\caption{The same four cells. The vertical gap within a line is what reasoning
buys; the gap between the lines is what the supervision buys. The lines are not
parallel, and they cross: that non-parallelism is the interaction, and its sign
says the two levers substitute.}
\label{fig:r4}
\end{figure}

\paragraph{We filled in the two missing cells, and supervision and reasoning turn
out to substitute.}
The previous version of this section said the comparison between
\RthreeTAcc\,\% and \LadderAlways\,\% was between two different checkpoints, so
the within-checkpoint value of reasoning after targeted supervision was unknown.
We then measured it. Both trained arms, all \RfourNSeeds\ seeds each, were
evaluated twice on the sealed benchmark --- reasoning off and reasoning on ---
through one serving harness, with the weights as the only thing that differed
(Table~\ref{tab:r4cells}, Figure~\ref{fig:r4}). The estimands, their intervals
and the rule for calling a result stable were fixed before the first of these
twelve evaluations ran.

With reasoning off, boundary-targeted supervision is ahead, as
Section~\ref{sec:r3} found on a different harness. With reasoning on, the two
arms are level or the random arm is slightly ahead. Turning reasoning on is worth
${+}\EoneDiff\,$pp on the targeted arm and ${+}\EtwoDiff\,$pp on the random arm,
and the difference between those two gains --- the interaction --- is
${-}\EfourDiff\,$pp with a 95\,\% interval of $[\EfourLo, \EfourHi]$ and the same
sign in all \RfourNSeeds\ seed pairs ($\EfourSeeds$). By the rule we registered,
that is \EfourStable.

So boundary supervision and test-time reasoning are \textbf{partial substitutes}
rather than independent sources of accuracy; the ledger's verdict for this
factorial is \RfourVerdict. Stated at the resolution the factorial actually
measures: the marginal value of reasoning is smaller after boundary-targeted
supervision than under the random-supervision control. This is an arm-level
estimate over the whole benchmark, not a per-boundary one, so we do not say that
reasoning is worth less on the particular boundaries the supervision covered
--- that would be a mechanism the design does not resolve. Targeted boundary data
earns its keep in a deployment that cannot afford to think; it does not stack on
top of reasoning. The reading we were careful not to assert before ---
``supervision and reasoning might be complementary'' --- is the one the data do
not support.

\begin{table}[t]
\centering
\begin{tabular}{lrr}
\toprule
Added supervision & No reasoning & Reasoning \\
\midrule
Random, budget-matched & 72.1\% & 88.7\% \\
Boundary-targeted & 75.2\% & 86.9\% \\
\bottomrule
\end{tabular}

\par\vspace{1em}
\begin{tabular}{llrlll}
\toprule
& Estimand & $\Delta$ (pp) & 95\% CI & Per seed & Verdict \\
\midrule
$E_1$ & Reasoning gain, targeted arm & ${+}11.7$ & $[8.1, 15.4]$ & +9.0, +13.0, +13.0 & stable \\
$E_2$ & Reasoning gain, random arm & ${+}16.6$ & $[12.1, 21.0]$ & +16.7, +15.0, +18.0 & stable \\
$E_3$ & Targeted ${-}$ random, reasoning on & ${-}1.8$ & $[-3.6, -0.1]$ & -4.0, +2.3, -3.7 & indeterminate \\
$E_4$ & Interaction & ${-}4.9$ & $[-8.1, -1.7]$ & -7.7, -2.0, -5.0 & stable \\
\bottomrule
\end{tabular}

\caption{Training $\times$ reasoning, all four cells measured on one harness,
\RfourNSeeds\ seeds per arm, \BlindN\ sealed items. Lower panel: the
pre-registered estimands. ``Stable'' requires both that the
\RfourNSeeds\ seed-paired differences share a sign and that the item-resampling
interval excludes zero; $E_3$'s interval excludes zero but its seed signs do not
agree, so the rule calls it \EthreeStable\ rather than letting us claim it. These
cells use the vLLM LoRA serving harness with its constrained-token probe, a
different harness from the pointer head of Table~\ref{tab:r3}; the reasoning-off
cells here are therefore not the numbers of Table~\ref{tab:r3} and should not be
compared with them.}
\label{tab:r4cells}
\end{table}

\paragraph{A ceiling would explain at most part of this.}
The obvious alternative to substitution is a ceiling: if the benchmark admits no
more than roughly \RfourCThink\,\% on exact action for this model, then both
reasoning-on cells are pressed against the same wall and the negative interaction
is an artefact of compression rather than a statement about the two levers. The
suspicion is reasonable --- no configuration we measured anywhere in this paper
exceeds \RfourCThink\,\% on exact action --- and the design cannot rule it out,
because establishing the ceiling would need a stronger arm we do not have.

What we can do is bound how much of the effect a ceiling could account for.
Pure compression toward a common ceiling leaves the better arm no worse than the
other, so it can shrink the reasoning-off gap of ${+}\RfourGapOff\,$pp toward zero
but cannot drive it negative. A ceiling can therefore account for at most
\RfourGapOff\,pp of the interaction's \EfourDiff\,pp magnitude, under a simple
common-ceiling compression model; a saturation that acts differently on the two
arms is not covered by that bound. The remaining
$\EthreeDiffSigned\,$pp requires the targeted arm to be genuinely \emph{worse}
once reasoning is on, which is $E_3$ --- and $E_3$ is \EthreeStable\ by our own
rule. So the honest accounting is that a ceiling explains at most \RfourGapOff\ of
the \EfourDiff\,pp, the residue points away from a ceiling, and we cannot
establish the residue. We therefore read
``partial substitutes'' as a description of the joint behaviour of the two budgets
in this operating range, not as a mechanism, and we say so again in
Section~\ref{sec:limitations}.

\paragraph{One caveat this does not remove.}
These four cells come from the serving harness and its constrained probe.
Section~\ref{sec:r3}'s \RthreeCAcc\,\% and \RthreeTAcc\,\% come from a different
decision mechanism, a trained pointer head, which cannot reason at all --- that is
why the missing cells could not simply be read off the earlier table. The two sets
of numbers are therefore not interchangeable, and we do not subtract across them.
Within this table the comparison is clean; across tables it is not.

\paragraph{What we can say.}
Supervision works on the boundaries you can name in advance and build contrastive
data for. On the boundary we deliberately withheld, we did not detect a gain. And no single-pass configuration we tried reached the
always-reasoning reference without reasoning --- the targeted arm sits
\LadderGap\,pp below it. But the factorial above says the
two budgets are not additive: where reasoning is available it absorbs most of what
targeted supervision buys. The operational reading is therefore narrower than
``spend on both''. Spend the data budget on boundaries you can name when the
deployment must answer in a single pass, and spend the compute budget when it can
afford to think. In this operating range, reasoning adds less after
boundary-targeted supervision than after the random-supervision control.

\section{Two robustness checks that need no new runs}
\label{sec:sensitivity}

Both checks below recompute from predictions we had already stored. Neither is a
new measurement and neither is pre-registered; they exist to show how much the
numbers above depend on two choices a reader cannot see from a table.

\paragraph{One metric name, three denominators.}
``\textsc{call\_tool} recall'' admits three defensible readings, and we have used
more than one of them. Counting only items whose gold action is exactly
\textsc{call\_tool} and requiring an exact match gives
${+}\RecallDefAPp\,$pp for targeted over random supervision. Widening the
denominator to every tool-requiring item, including
\textsc{call\_multiple\_tools}, while still requiring the exact action gives
${+}\RecallDefBPp\,$pp --- this is the definition our pre-registration gate
actually scored. Widening further so that any tool call counts as correct on a
tool-requiring item gives ${+}\RecallDefCPp\,$pp. The same effect is therefore
anywhere from ${+}\RecallDefCPp$ to ${+}\RecallDefAPp\,$pp, a factor of
\RecallDefSpread, depending only on how the metric is named. We now write
\emph{\textsc{call\_tool}-only recall} ($n=\NUnitsTool$) and \emph{any-tool
recall} ($n=\NUnitsToolRthree$) and give both, because the one endpoint that
cleared our bar is the one this choice moves most.

\paragraph{Dropping the two layers where a lexical shortcut exists.}
Section~\ref{sec:bench} reports that two of the \ShortcutLayers\ layers can be
beaten from the request alone. Recomputing every main effect on the
\CleanLayers\ layers that passed, \CleanNItems\ items, gives
Table~\ref{tab:sensitivity}. Nothing we claim gets weaker: the reasoning effect
grows from ${+}\SensFullReason$ to ${+}\SensCleanReason\,$pp and the supervision
effect from ${+}\SensFullTminusC$ to ${+}\SensCleanTminusC\,$pp. One number does
change character --- the reasoning-off format effect moves from exactly
$\SensFullFormatOff$ to ${-}\SensCleanFormatOffAbs\,$pp, so on the clean subset generated JSON is
nominally ahead rather than exactly level. Both values are far below what this
sample can resolve, which is the point: that cell was never evidence of a
direction, and the clean subset makes that visible instead of hiding it behind a
tidy zero.

\begin{table}[t]
\centering
\begin{tabular}{lrr}
\toprule
Effect (pp) & All 300 items & Shortcut-clean 240 \\
\midrule
Reasoning effect (single token) & ${+}17.3$ & ${+}20.8$ \\
Format effect, reasoning off & $0.0$ & ${-}0.8$ \\
Format effect, reasoning on & ${+}2.7$ & ${+}4.2$ \\
Targeted ${-}$ random, exact action & ${+}4.8$ & ${+}5.4$ \\
Targeted ${-}$ random, CALL\_TOOL-only recall & ${+}15.7$ & ${+}17.8$ \\
Targeted ${-}$ random, any-tool recall & ${+}9.9$ & ${+}11.7$ \\
\bottomrule
\end{tabular}

\caption{Every main effect recomputed on the \CleanLayers\ layers whose lexical
shortcut probes stayed under their ceiling. Recomputed from stored predictions;
no new runs, and not pre-registered. Positive favours reasoning on, the single
token, or targeted supervision respectively.}
\label{tab:sensitivity}
\end{table}

\section{Zero-shot transfer across policy languages}
\label{sec:english}

Every enterprise and boundary-targeted record this model was trained on is
Korean. We asked whether the decision behaviour measured above depends on that,
and pre-registered the audit, its gates and its stopping rule before rendering a
single English item.

\paragraph{Design.} The benchmark's labels are computed from latent factors rather
than read off text, so the same \REnRenderTotal\ latent scenarios can be rendered
again in another language without touching a single label. We did not translate
the Korean items, which would have carried Korean surface cues into English; each
scenario was rendered afresh in English from its fact specification, by the same
renderer that produced its Korean twin, with amounts, identifiers and proper nouns
left verbatim so that only the language of the prose changes. English items pass
the same gate axes as Korean ones --- leakage, external references, withheld and
required values, factor recovery by two independent extractors, and an adversarial
challenge. We crossed the language of the policy document with the language of
the surrounding instructions, and rendered every scenario a second time in Korean
to measure how much a fresh draw of the renderer moves accuracy on its own. All
renderings, both reasoning settings and both arms of Section~\ref{sec:ladder} were
evaluated on one server in one run, over the \REnN\ items that passed the gates in
every rendering.

That common subset is easier than the full benchmark: the Korean rendering scores
${+}\REnSelOff$ pp higher without reasoning and ${+}\REnSelOn$ pp higher with it
than it does on all \REnRenderTotal\ items. The absolute accuracies in
Table~\ref{tab:english} are therefore not comparable with the main tables; only
the contrasts within it are.

\paragraph{No English degradation at this resolution.} With reasoning on, the
English rendering scores \REnEnOn\,\% against \REnKoOn\,\% for Korean, a
difference of ${-}\REnXtwo$ pp $[\REnXtwoLo, {+}\REnXtwoHi]$ that \REnGateTwo\ the
pre-registered retention gate (no worse than $\REnFloorTol$ pp, interval above
$\REnCiTol$ pp). Without reasoning the difference is ${+}\REnXone$ pp.
Re-rendering the same scenarios in Korean moves accuracy by ${-}\REnXthree$ pp
with reasoning and ${-}\REnXthreeb$ pp without, so the magnitude of the observed
language difference is comparable to the variation introduced by re-rendering the
same latent scenarios in Korean. Our pre-registered rule for calling a language
effect at all asks for twice that floor, and it is not met. We therefore read the
result as no detectable English degradation within this audit, not as evidence
that the two languages are equivalent. Nor is the small gap an artefact of
offsetting errors: on the same items the two renderings lead to the same action
\REnAgree\,\% of the time and are both correct on \REnBoth\,\%.

Crossing frame and document language decomposes the small gap descriptively.
Holding the frame fixed, English documents score ${-}\REnDocKoFr$ pp
$[\REnDocKoFrLo, {+}\REnDocKoFrHi]$ under a Korean frame and ${-}\REnDocEnFr$ pp
$[\REnDocEnFrLo, {+}\REnDocEnFrHi]$ under an English one; holding the document
fixed, an English frame moves the score by ${+}\REnXseven$ pp on Korean documents
and by \REnXsix\ pp on English ones. What gap there is sits with the document
rather than the instructions, and none of these intervals excludes zero. These
contrasts were not pre-registered. Among the unplanned contrasts one does exclude
zero --- an English frame on English documents without reasoning, ${+}\REnFrEnDocOff$ pp
$[{+}\REnFrEnDocOffLo, {+}\REnFrEnDocOffHi]$ --- and with several such contrasts
examined we do not read it as an effect.

\paragraph{Does the decision behaviour transfer?} Similar accuracy alone
would be consistent with English ability simply having survived Korean
fine-tuning. The factorial of Section~\ref{sec:ladder} gives a sharper test,
because it has a signature: targeted supervision helps when the model cannot
reason, and that help disappears when it can. Without reasoning, targeted over
budget-matched supervision is worth ${+}\REnTCKoOff$ pp
$[\REnTCKoOffLo, {+}\REnTCKoOffHi]$ in Korean and ${+}\REnTCEnOff$ pp
$[{+}\REnTCEnOffLo, {+}\REnTCEnOffHi]$ in English; with reasoning it is
${-}\REnTCKoOn$ and \REnTCEnOn\ pp, neither interval excluding zero. By the
registered stability rule the reasoning-off gain is \REnTCKoOffStable\ in Korean
(its item-clustered interval reaches zero) and \REnTCEnOffStable\ in English (one
seed pair changes sign, \REnTCEnOffSeeds\,pp). The point estimates have the same
shape in both languages, but we did not detect a stable targeted-supervision gain
in either, so we do not claim that this signature transferred; what the audit
supports is the absence of a detectable English degradation.

\paragraph{Gates.} The retention gate and the reasoning-off non-regression gate
\REnGateFour. The per-boundary gate \REnGateThree. All \REnBndMagOk\ of its
\REnBndN\ magnitude conditions held, but the rule also required the seed-paired
sign to agree within each boundary, and a difference near zero is expected to
change sign across seeds. We regard that requirement as a design error in a
non-regression gate and do not alter the gate after the fact; the claim of this
section rests on the retention gate alone.

\begin{table}[t]
\centering
\begin{tabular}{llrr}
\toprule
Frame & Document & No reasoning & Reasoning \\
\midrule
Korean & Korean & 77.51\% & 89.16\% \\
Korean & English & 76.34\% & 88.34\% \\
English & Korean & 76.81\% & 89.63\% \\
English & English & 77.86\% & 88.34\% \\
\midrule
Korean & Korean, re-rendered & 76.22\% & 88.69\% \\
\bottomrule
\end{tabular}

\vspace{0.6em}

\begin{tabular}{lll}
\toprule
Contrast (pp) & No reasoning & Reasoning \\
\midrule
English ${-}$ Korean & ${+}0.35$ $[-2.45, 3.03]$ & ${-}0.82$ $[-2.68, 0.93]$ \\
Korean re-render ${-}$ Korean & ${-}1.28$ $[-3.38, 0.82]$ & ${-}0.47$ $[-1.63, 0.58]$ \\
Targeted ${-}$ random, Korean & ${+}3.03$ $[0.00, 6.18]$ & ${-}0.93$ $[-2.45, 0.58]$ \\
Targeted ${-}$ random, English & ${+}2.91$ $[0.58, 5.36]$ & $0.00$ $[-1.63, 1.63]$ \\
\bottomrule
\end{tabular}

\caption{Zero-shot transfer to English. All cells use the \REnN\ items valid in
every rendering; this subset is \REnSelOn--\REnSelOff\,pp easier than the
full benchmark, so absolute accuracies are not comparable with other tables.
Top: exact action under each combination of instruction-frame and
policy-document language, and a second independent Korean rendering. Bottom:
paired contrasts with item-bootstrap 95\% intervals, means over \RfourNSeeds\ training
seeds of the boundary-targeted arm (first two rows) or the targeted and
budget-matched arms (last two). No English data was used in training.}
\label{tab:english}
\end{table}

\section{An open re-rendering as a robustness check}
\label{sec:openrerender}

The benchmark's surface text was written by two commercial models
(Section~\ref{sec:bench}). We cannot redistribute that text without carrying their
providers' terms with it, so the original rendering stays private and remains the
confirmatory benchmark for every primary analysis in this paper. For public release
we re-rendered the same latent scenarios, with the same executable gold labels and
proofs, using two open-weight models, Qwen3.6-27B and Gemma-4-31B. A hash of the
scenario assigned the renderer, so counterfactual twins and the two languages of a
scenario always share one. The same gates were applied, with commercial models
acting only as judges: factor recovery by two extractors and a challenger. Items
that failed received one retry with the same renderer. The five rendering
configurations of the cross-lingual audit --- Korean, English, the two mixed
frame/document variants, and the second Korean rendering --- were then sealed on
their common intersection of \OpenN\ items, and every analysis was repeated on them without changing a character
of the sealed text.

One fact limits what this can show. The re-rendering was generated after the
results on the original rendering were known. No model under evaluation took part
in rendering or selection, and the gates never see predictions, but the benchmark
surface is still a post-hoc object. We therefore read it as a robustness check on
the original findings, not as an independent confirmation of them.

The selection is not neutral in difficulty. The gates rejected \OpenNDropped\ of the
\BlindN\ scenarios in at least one rendering, and they did so unevenly: only
\OpenPassMulti\,\% of multi-tool scenarios and \OpenPassClarify\,\% of clarification
scenarios survived, against roughly nine in ten elsewhere. Measured with the
original rendering's own predictions --- a post-hoc diagnostic, never an input to
selection --- the rejected scenarios are harder by \OpenKeptDropped\,pp
$[\OpenKeptDroppedLo, \OpenKeptDroppedHi]$, and the retained \OpenN\ are
\OpenEasier\,pp easier than the full sealed set (\OpenEasierLo\ to \OpenEasierHi\,pp
across sixteen cells). Absolute accuracies on the two renderings are therefore not
comparable, and we compare only paired contrasts measured within each rendering.
Latency is excluded too: the re-measurement shared servers across cells, so its
queueing conditions differ from those of Section~\ref{sec:indist}.

The grades in Table~\ref{tab:replication} are a post-hoc, descriptive taxonomy, not
pre-registered endpoints. Each finding is graded by a rule written in code:
\emph{replicated} when both renderings resolve the effect in the same direction,
\emph{null replicated} when neither resolves it, \emph{directionally consistent}
when only one resolves it and the directions agree, and \emph{did not replicate}
otherwise; a categorical verdict whose literal label differs is reported as such
and not graded as a replication. We revised the rule once after a first pass,
so that the factorial and label-matched estimands use the stability criterion their
pre-registration defines (same sign in every seed and an interval excluding zero)
rather than the interval alone, and so that a null that became an effect is
not counted as consistent. The revision changed four rows. Three became stricter: the two
label-matched contrasts moved from ``replicated'' to ``directionally consistent'',
and the reasoning-off English contrast from ``directionally consistent'' to ``did
not replicate''. One became more lenient: $E_3$ moved from ``did not replicate'' to
``null replicated'', because its seeds disagree in sign on the original
rendering, which is why Section~\ref{sec:ladder} already reported it as
indeterminate. Of the \OpenRows\ graded findings, of which Table~\ref{tab:replication}
shows the main ones, the first-pass rule gave \OpenVzeroRepl\ replicated,
\OpenVzeroNull\ null replicated, \OpenVzeroDir\ directionally consistent and
\OpenVzeroNot\ not replicated; the final rule gives \OpenRepl, \OpenReplNull,
\OpenDirCons\ and \OpenNotRepl, with \OpenLabel\ verdict whose literal label
changed. We report both so that the revision can be checked.

\begin{table}[t]
\centering
\small
\begin{tabular}{@{}>{\raggedright\arraybackslash}p{4.2cm}ll>{\raggedright\arraybackslash}p{2.6cm}@{}}
\toprule
Finding (pp) & Original (sealed) & Open re-rendering & Grade \\
\midrule
Reasoning, single-token column & ${+}17.3$ $[12.3, 23.0]$ & ${+}21.8$ $[16.1, 27.8]$ & Replicated \\
Format, reasoning off & ${+}0.0$ $[-4.3, 4.0]$ & ${-}4.8$ $[-10.1, 0.4]$ & Null replicated \\
Format, reasoning off, \textsc{call\_tool}-only recall ($n=89$) & ${+}12.4$ $[4.5, 20.2]$ & ${+}2.6$ $[-5.3, 10.5]$ & Did not replicate \\
Format, reasoning on & ${+}2.7$ $[-0.7, 6.0]$ & ${+}2.4$ $[0.0, 4.8]$ & Null replicated \\
Targeted ${-}$ random, exact action & ${+}4.8$ (lo $+2.0$) & ${+}7.7$ (lo $+4.7$) & Replicated \\
Targeted ${-}$ random, CF both-correct & ${+}8.9$ (lo $-1.1$) & ${+}13.9$ (lo $+2.8$) & Directionally consistent \\
Targeted ${-}$ random, any-tool recall ($n=118$) & ${+}9.9$ (lo $+4.2$) & ${+}14.9$ (lo $+8.8$) & Replicated \\
$E_1$ reasoning gain, targeted & ${+}11.7$ $[8.1, 15.4]$ & ${+}14.1$ $[9.8, 18.6]$ & Replicated \\
$E_2$ reasoning gain, random & ${+}16.6$ $[12.1, 21.0]$ & ${+}17.5$ $[12.5, 22.7]$ & Replicated \\
$E_3$ targeted ${-}$ random, reasoning on & ${-}1.8$ $[-3.6, -0.1]$ & ${-}0.4$ $[-1.8, 1.1]$ & Null replicated \\
$E_4$ interaction & ${-}4.9$ $[-8.1, -1.7]$ & ${-}3.4$ $[-6.8, 0.1]$ & Did not replicate \\
Targeted ${-}$ label-matched (C1) & ${+}5.8$ $[2.7, 8.9]$ & ${+}8.3$ $[4.6, 12.2]$ & Directionally consistent \\
Single token ${-}$ pointer, reasoning off & ${-}3.9$ $[-7.6, -0.2]$ & ${-}7.0$ $[-11.3, -3.1]$ & Replicated \\
English ${-}$ Korean, reasoning on & ${-}0.8$ $[-2.7, 0.9]$ & ${-}0.3$ $[-1.8, 1.1]$ & Null replicated \\
English ${-}$ Korean, reasoning off & ${+}0.3$ $[-2.5, 3.0]$ & ${+}4.6$ $[1.5, 7.7]$ & Did not replicate (null $\to$ effect) \\
\addlinespace
R4 pre-registered verdict (literal) & substitution & covers supervision gap & Literal label changed \\
R6 transfer verdict, four gates & transfers & transfers, same gates & Replicated \\
Matched-format branches (P4) & collapsed & collapsed & Replicated \\
\bottomrule
\end{tabular}

\caption{Replication on the open re-rendering. \emph{Original}: the sealed, private
rendering on which every primary analysis rests ($n=\BlindN$; \REnN\ for the
English rows). \emph{Open}: the public re-rendering of the same latent scenarios and
gold labels with open-weight models, sealed on \OpenN\ common items and generated
after the original results were known. Paired contrasts in pp with item-bootstrap
95\% intervals (lower bound only for the pre-registered endpoints of
Section~\ref{sec:r3}); format contrasts are single token minus JSON. Grades use the post-hoc robustness taxonomy of Section~\ref{sec:openrerender};
they are descriptive, assigned by code from the ledger, and are not pre-registered
endpoints. R4 and C1 rows use the pre-registered stability rule.}
\label{tab:replication}
\end{table}

Three findings replicated. On the open rendering, the reasoning effect in the
single-token column is ${+}\OpenThinkDiff\,$pp $[{+}\OpenThinkLo, {+}\OpenThinkHi]$;
the English transfer verdict and all four of its gates are unchanged; and the
matched-format branches collapsed again. The training-by-reasoning interaction,
which is stable on the sealed set (Section~\ref{sec:ladder}), did not replicate
under item-clustered intervals: on the open rendering it is
${-}\OpenEfourDiff\,$pp $[\OpenEfourLo, {+}\OpenEfourHi]$, which the registered rule
calls \OpenEfourStable\ and Table~\ref{tab:replication} grades
``\OpenEfourGrade''. The point estimate keeps the negative sign, but the interval
reaches zero. The literal pre-registered verdict of Section~\ref{sec:ladder} did not
replicate: the rule returns ``reasoning covers the supervision gap'' rather than
``substitution''. The difference arose from decision-rule precedence. On the open
rendering the two reasoning-on arms land within one point of each other, and the
rule checks that condition before it checks the interaction. So neither the
literal verdict nor the stability of the interaction that our C4 interpretation
rests on replicated; C4 rests on the sealed rendering alone.

Two further findings did not replicate. The single-token advantage on \textsc{call\_tool}-only recall
($n=\NUnitsTool$) with reasoning off (Section~\ref{sec:twobytwo}) falls from ${+}\XFormatToolDiff$ to
${+}\OpenFmtToolDiff\,$pp $[\OpenFmtToolLo, {+}\OpenFmtToolHi]$. The subset does not
explain it: restricting the original rendering to the same \OpenN\ items leaves the
effect at ${+}\OpenOrigFmtToolDiff\,$pp $[{+}\OpenOrigFmtToolLo, {+}\OpenOrigFmtToolHi]$.
It belonged to the surface text. The reasoning-off format contrast on exact action
moved the other way, to $\OpenFmtOffDiff\,$pp $[\OpenFmtOffLo, {+}\OpenFmtOffHi]$,
still an interval that includes zero. Read together, they give C2's qualifier its
stronger form: we find no stable evidence that the action-output format changes
decision accuracy, and the one format effect we resolved was a property of one
rendering. The other non-replication is the English-minus-Korean contrast with
reasoning off, which became ${+}\OpenEnKoOff\,$pp. The re-rendering noise floor
moved with it. English against the second Korean rendering is
${+}\OpenEnKoTwo\,$pp $[\OpenEnKoTwoLo, {+}\OpenEnKoTwoHi]$, an interval that
includes zero, so we do not read it as an English advantage; the pre-registered
transfer gate is judged as originally defined.

The shortcut probes of Section~\ref{sec:bench} were themselves sensitive to the
rendering. The dialect shortcut persisted and grew stronger: on the open rendering,
the policy text or the request alone recovers every dialect item. The
missing-information shortcut disappeared, leaving six clean layers instead of five.
The label-matched control of Section~\ref{sec:posthoc} became seed-consistent on
the open rendering; because that rendering came after the original result, we do
not use it to strengthen C3's qualifier.

\section{Limitations}
\label{sec:limitations}

Five limitations bound every claim in this paper. The remaining caveats, which
matter for auditing particular numbers rather than for reading the claims, are
collected in Appendix~\ref{app:caveats}.

\paragraph{One model and one constructed policy family.} Every result comes from
one 27B Korean enterprise model and one family of constructed policies. The sealed
benchmark is computed rather than harvested, which is what makes its labels
defensible and also what limits its ecological validity: a model could do well on
these policies and badly on a real company's documents. Evaluation on real
enterprise policies, with labels their owners can defend, is the most important
next validation, and this paper does not provide it.

\paragraph{The reasoning-on output-format comparison is confounded.} The reasoning
prompt of the single-token arm is not that of the JSON arm: the first design put
only the deployed instruction in stage one and no action code appeared in the
top-20 logits, so stage one was changed to carry the same user content plus the
single-token suffix. \emph{The reasoning in the single-token arm is therefore not
the same reasoning as under the JSON contract}, and the reasoning row of the
two-by-two carries a prompt difference alongside the format difference. The
post-hoc control of Section~\ref{sec:posthoc} tried to remove it and did not,
because the output instruction changed the reasoning trace itself, and a
pre-registered matched-state diagnostic that froze the trace was void because its
positive control failed (Appendix~\ref{app:matchedstate}). The
reasoning-on format effect is therefore unresolved; only the reasoning-off format
contrast is cleanly identified. The in-distribution comparison of
Section~\ref{sec:indist} has the same problem, since its two paths differ in
reasoning as well as format, and we did not repeat its restart measurement under
matched reasoning.

\paragraph{The targeted-versus-random comparison shifts the class prior.} The treatment and control arms were matched on token budget, not on label
distribution. The added treatment records contain \textsc{call\_tool} in \PriorTreatTool\,\%
of cases against \PriorCtrlTool\,\% for the control --- a factor of
\PriorToolRatio\ --- and the benchmark's own gold distribution puts
tool-requiring items at \GoldToolShare\,\% of the set. The single endpoint that cleared the pre-registered bar is
any-tool recall. A class-prior shift is therefore a live alternative
explanation for that result and for part of the overall gain, and our design does
not exclude it: the pre-registered analysis reported neither the treatment arm's
predicted action distribution nor its \textsc{call\_tool} precision, and the
pre-registered design did not include a label-matched third arm. Both gaps are
filled only post hoc. Section~\ref{sec:posthoc} adds the label-matched arm; it
argues against the class prior as a sufficient explanation without conclusively
rejecting it. The same section recomputes per-class precision, recall and the
predicted action distribution from the stored predictions
(Table~\ref{tab:c0c1t}, Figure~\ref{fig:c0c1t}). It shows the targeted arm's
recall and predicted \textsc{call\_tool} rate rising together at roughly constant
precision, which cannot separate better boundary detection from a shift toward
predicting \textsc{call\_tool}.

\paragraph{Seed and retraining uncertainty is not in the intervals.} The boundary-targeted comparison's intervals resample items while holding the
\RthreeNSeeds\ trained seed pairs fixed, so they do not carry retraining
uncertainty. The per-seed deltas are wide enough that this matters --- overall
\RthreeSeedsAcc\,pp, counterfactual \RthreeSeedsCf\,pp, any-tool recall
\RthreeSeedsTool\,pp --- and we report them alongside every interval rather than
asking the reader to infer them. With \RthreeNSeeds\ seeds we do not think a
hierarchical interval would be stable enough to be worth quoting. The two-by-two cells are single runs. Reasoning with a generated action object is
not reproducible at temperature zero in our setting --- a rerun changed a
measurable fraction of the \BlindN\ decisions --- so those cells carry run-level
noise that the intervals, which resample items, do not capture. The cell that
reasons and then reads one token is in the same position for the same reason: its
reasoning phase is autoregressive, and we did not establish whether it shares the
restart variability we measured for the deployed path. Only the no-reasoning
cells are free of this.

\paragraph{The public re-rendering is post hoc and selected.} The public re-rendering (Section~\ref{sec:openrerender}) was generated after the
original results were known, so it is a robustness check rather than an independent
confirmation. Its gates also removed hard multi-tool and clarification scenarios
disproportionately: the benchmark we release is easier than the one our primary
claims rest on, by \OpenEasier\,pp on the original rendering's own predictions.

\section{Related work}

\paragraph{Generating structure versus constraining it.}
Making a language model emit a well-formed structure is usually treated as a
decoding problem: finite-state or grammar-constrained decoding masks the
vocabulary so that only valid continuations remain
\citep{willard2023outlines,geng2023gcd}. The line runs from incremental parsing
for formal languages \citep{scholak2021picard} through grammar- and
schema-constrained decoding at production scale
\citep{dong2025xgrammar,geng2025jsonschemabench}, and recently to evidence that
structural constraints can change task accuracy rather than only syntactic
validity \citep{schall2025hiddencost}. These works ask how to generate a valid
structure efficiently. Our question is upstream of that. When the structure to be produced is
one of seven known actions, the generation step is not merely constrainable, it
is removable, and Section~\ref{sec:indist} measures what removing it costs and
saves.

\paragraph{Task-oriented dialogue.}
It is tempting to read our in-distribution result as a return to modular
intent-and-slot pipelines, and that reading would be wrong. The field moved the
other way for good reasons: a single generative model over belief state, action
and response outperformed modular pipelines on multi-domain benchmarks
\citep{hosseiniasl2020simpletod,peng2021soloist}. Our claim is narrower than a
reversal. We hold the model and its training fixed and remove only the
autoregressive emission of the action symbol, in a setting where a downstream
system executes the action; the generation step is not what produces the accuracy
there.

\paragraph{Reasoning before action.}
Chain-of-thought prompting and self-consistency established that explicit
intermediate computation can materially improve a final decision
\citep{wei2022cot,wang2023selfconsistency}, and \textsc{ReAct} interleaves that
reasoning with actions taken in an external environment \citep{yao2023react}.
More recent reasoning models and test-time scaling methods make inference compute
an explicit budget \citep{snell2025scaling,muennighoff2025s1,deepseek2025r1}. Our
factorial differs in treating training supervision and test-time reasoning as two
separately measurable resources spent on the same bounded policy decision.

\paragraph{Learning versus reasoning.}
\textsc{STaR} and work descended from it convert generated rationales into
training signal, moving computation from inference time back into the weights
\citep{zelikman2022star}; test-time scaling asks the opposite question of how
much computation to spend after training is over \citep{snell2025scaling}. Our
training ${\times}$ reasoning factorial places those two budgets in one
controlled decision task and estimates their interaction, rather than evaluating
either resource on its own.

\paragraph{Tool use and policy-governed agents.}
Tool selection and API invocation are by now core language-model capabilities
\citep{schick2023toolformer,patil2024gorilla,qin2024toolllm,li2023apibank}. More
recent benchmarks move from isolated function calls toward stateful,
policy-governed interaction: \textsc{BFCL} evaluates abstention alongside serial
and parallel calls \citep{patil2025bfcl}, \textsc{ToolSandbox} adds state
dependencies and deliberately insufficient information \citep{lu2025toolsandbox},
and $\tau$-bench evaluates agents that must use domain APIs while obeying written
policy rules in conversation with a user \citep{yao2025taubench}.
\textsc{WorkArena} supplies the complementary enterprise-software setting
\citep{drouin2024workarena}. Our seven-way decision isolates the action-selection
boundary that these systems normally leave embedded inside a longer trajectory.

\paragraph{Deciding after reasoning rather than generating the decision.}
Reading a decision off a hidden state after a reasoning pass, rather than
generating it, has recently appeared in deployed systems: Jeeves-9B attaches a
pointer head to a 9B model and reports a gain from no-think to think of a similar
shape to our reasoning effect \citep{waltz2026jeeves}. Relatedly, the final answer
of a reasoning model is decodable from intermediate activations well before the
trace ends, which supports probe-based early exit \citep{boppana2026reasoningtheater}.
Our pointer-head arm is a diagnostic rather than a proposed method, and we report
it as such, including its calibration cost.

\paragraph{When to spend test-time computation.}
Adaptive-computation work asks the complementary question of whether to reason at
all, through reinforcement learning over a think/no-think choice
\citep{zhang2025adaptthink,tu2025autothink}, over discrete budget modes
\citep{kassenaar2026whentothink}, and through stopping criteria computed from the
trace itself \citep{li2026trace}. A finding we reproduce in a different setting is
that a single confidence-like signal does not map cleanly onto whether more
computation would help; \citet{li2026dial} show the same signal can predict both
benefit and harm depending on the environment, and recommend learning the
direction rather than assuming it. Our confidence router hits the same wall from
the other side: on unseen boundaries the non-reasoning model is often confident
and wrong.

\paragraph{Benchmarks whose labels are computed.}
Building evaluation sets whose ground truth comes from a simulator or a seeded
script, with language models restricted to producing surface text, has been used
for insider-threat detection \citep{flynt2026orgforge} and for agent memory
\citep{spencer2026groundtruthfirst}. The same principle underlies
\textsc{WebArena} and \textsc{SWE-bench}, where success is read from environment
state or from executable tests rather than from an evaluator's free-form
judgement \citep{zhou2024webarena,jimenez2024swebench}, and $\tau$-bench and
\textsc{ToolSandbox} apply it to tool-using interaction
\citep{yao2025taubench,lu2025toolsandbox}. Section~\ref{sec:bench} applies the same
discipline to policy decisions, and adds two things that our failure modes forced
on us: a proof emitted alongside each label, and counterfactual pairs in which the
single changed variable is known by construction rather than asserted after the
fact.

\paragraph{Cross-lingual instruction transfer.}
Instruction-following transfers across languages to a substantial degree without
per-language supervision, through multitask finetuning over multilingual
pretrained representations \citep{muennighoff2023crosslingual}, in instruction
tuning specifically \citep{chirkova2024zeroshot}, from only a small fraction of
multilingual data \citep{shaham2024pinch}, and at scale
\citep{ustun2024aya}. Our English audit is narrower than any of these. It does
not establish multilingual capability; it tests whether one decision behaviour,
learned entirely from Korean enterprise supervision, survives an independent
English rendering of the same executable policy scenarios.

\paragraph{What we believe is new.}
The closest prior work evaluates tool selection, policy adherence, structured
generation or adaptive reasoning separately. We are not aware of work that
factorially separates the action-output interface from test-time reasoning for a
bounded enterprise-policy decision, and then measures the interaction between
boundary-targeted supervision and test-time reasoning on the same sealed,
executable benchmark.

\section{Two post-hoc experiments, one of which failed}
\label{sec:posthoc}

Neither experiment below is pre-registered. Both were designed after we had seen
the sealed benchmark, and we name them post-hoc so that no reader mistakes them
for confirmatory evidence. Section~\ref{sec:r3}'s pre-registered status is
unchanged by either.

\paragraph{A label-matched control: matching the class prior does not eliminate
the average gain.}
The sharpest objection to Section~\ref{sec:r3} is that the treatment arm did not
merely receive better boundary examples, it received \textsc{call\_tool} at
\PriorToolRatio$\times$ the control's rate. We therefore built a second control,
C1, whose 7-action histogram matches the treatment arm \emph{exactly} and which
contains no counterfactual pairing or boundary contrast --- the same labels in the
same proportions, learned from data that never puts two decisions side by side.
Three seeds, same recipe, same schedule (Table~\ref{tab:c1}).

Matching the prior did not absorb the effect in two of the three seeds. The
targeted arm's advantage over the label-matched control, ${+}\TmConeDiff\,$pp
averaged over seeds, is if anything larger than its advantage over the original
control, ${+}\TmCzeroDiff\,$pp, and the label-matched control did not beat the
original one ($\ConemCzeroDiffSigned\,$pp). That argues against the class prior
as a \emph{sufficient} explanation of Section~\ref{sec:r3}. It does not reject
the class-prior hypothesis outright: the third seed ($\TmConeSeeds$) is the case
in which matching the prior did absorb the advantage, and one seed out of three is
not a result we can wave away.

It also does not establish the alternative. One of the three seed pairs flips
sign ($\TmConeSeeds$), so by the stability rule we fixed before looking, $T-C1$ is
\TmConeStable. The label-matched arm is also markedly noisier across seeds, a
\ConeSpread\,pp range against the original control's \CzeroSpread. Two statements have to be kept apart here, and neither is fully settled.
Matching the control arm's action-label distribution did not eliminate the
targeted arm's advantage on average, which argues against the class prior as a
\emph{sufficient} explanation of Section~\ref{sec:r3} --- but seed-level
instability stops that from being a conclusive rejection of the class-prior
hypothesis. And the same instability prevents us from isolating contrastive
boundary construction itself as the causal mechanism. That is weaker than we
hoped for and stronger than the objection.

\begin{table}[t]
\centering
\begin{tabular}{lrrr}
\toprule
Added supervision & Exact action & Boundary layers ($n=120$) & CALL\_TOOL-only \\
\midrule
Random, budget-matched (C0) & 74.4\% & 73.6\% & 59.2\% \\
Random, \textbf{label-matched} (C1) & 73.4\% & 73.1\% & 56.2\% \\
Boundary-targeted (T) & 79.2\% & 79.4\% & 74.9\% \\
\addlinespace
\quad $T-C1$ exact action & ${+}5.8$ & $[2.7, 8.9]$ & indeterminate \\
\quad $T-C1$ boundary layers & ${+}6.4$ & $[0.6, 12.5]$ & indeterminate \\
\quad $T-C0$ exact action & ${+}4.8$ & $[2.0, 7.7]$ & stable \\
\quad $C1-C0$ exact action & ${-}1.0$ & $[-3.1, 1.1]$ & indeterminate \\
\bottomrule
\end{tabular}

\caption{The label-matched control (post-hoc, not pre-registered). Upper panel:
arm means over \ConeNSeeds\ seeds on the sealed benchmark. C0 is the same
checkpoints as the random arm of Table~\ref{tab:r3}, so its exact-action value
matches that table. \emph{Boundary layers} is item accuracy over all
$n=\ConeBndN$ items of the \ConeBndNLayers\ boundary-constructed layers (answer
versus tool, confirm versus refuse, missing information), whatever their gold
action, averaged over seeds; it excludes the policy-versus-principle layer. It is
therefore not the pooled boundary groups of Table~\ref{tab:r3} ($n=\BndNPool$),
which select items by layer \emph{and} gold-action pair and include the held-out
boundary in both of its layers. Lower panel:
contrasts, their item-resampling intervals, and the stability verdict. C1 matches
the treatment arm's action-count histogram exactly --- the builder asserts
equality and refuses otherwise --- and withholds the contrastive pairing. The
separate \emph{token} budget lands within \BudgetSkewPct\,\% overall, with two
actions short of their per-label token target because the factory pool contains
no records long enough; the counts are exact, the token totals are not. The
last column of the upper panel is \textsc{call\_tool}-only recall over the
$n=\NUnitsTool$ items whose gold action is \textsc{call\_tool}, not the any-tool
recall of Table~\ref{tab:r3} ($n=\NUnitsToolRthree$).}
\label{tab:c1}
\end{table}

\begin{figure}[t]
\centering
\includegraphics[width=0.49\linewidth]{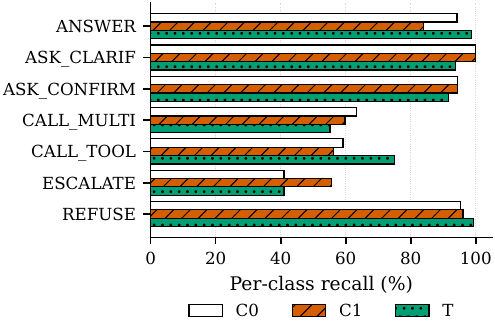}\hfill
\includegraphics[width=0.49\linewidth]{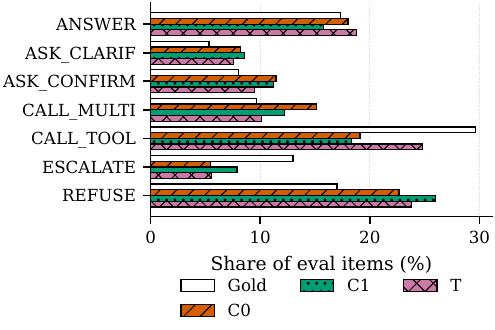}
\caption{Per-class recall (left) and the marginal predicted-action distribution
against the gold distribution (right) for C0, C1 and T, seed means over \CctNSeeds\
seeds on the \CctNItems\ sealed items. The right panel is a marginal, not a
confusion matrix; the full matrix is Figure~\ref{fig:c0c1tcm}. T's \textsc{call\_tool} recall and predicted
\textsc{call\_tool} share rise together; its \textsc{call\_tool} precision does not
(Table~\ref{tab:c0c1t}). The two panels are separate figures and assign colours
to arms independently; read each against its own legend.}
\label{fig:c0c1t}
\end{figure}

\begin{figure}[t]
\centering
\includegraphics[width=\linewidth]{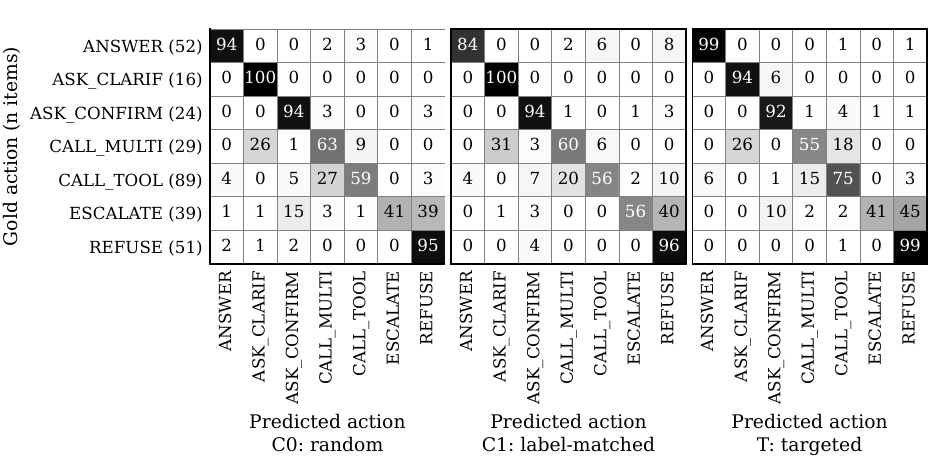}
\caption{Full $7{\times}7$ confusion matrices for C0, C1 and T on the \CctNItems\
sealed items (post-hoc, stored pointer-head predictions). Rows are the gold action
with its item count, columns the predicted action. Each cell is the seed-mean count
over \CctNSeeds\ seeds divided by the row's gold count, in \%; the grey level
encodes the same number, which is printed so the figure reads without colour. Rows
sum to 100 up to rounding. Per-action precision and recall are in
Table~\ref{tab:c0c1tperclass}.}
\label{fig:c0c1tcm}
\end{figure}

\paragraph{Per-class view of the targeted advantage.}
To see whether the targeted arm's advantage is a \textsc{call\_tool} prior effect,
we recomputed per-class metrics from the stored pointer-head predictions of C0, C1
and T (\CctNSeeds\ seeds each, \CctNItems\ items, \CctGoldN\ with gold
\textsc{call\_tool}; Table~\ref{tab:c0c1t}, Figure~\ref{fig:c0c1t}). This is a
post-hoc recomputation and was not pre-registered. The full confusion matrices and
per-action precision and recall for all seven actions are in
Figure~\ref{fig:c0c1tcm} and Table~\ref{tab:c0c1tperclass}. Seed-mean \textsc{call\_tool}
precision was \CctCzeroPrec\,\% for C0, \CctConePrec\,\% for C1 and
\CctTPrec\,\% for T, so T did not gain precision, and its seed-level precision
differences were mixed in sign (T$-$C0: \CctTCzeroPrecSeeds\,pp; T$-$C1:
\CctTConePrecSeeds\,pp). The gain was in \textsc{call\_tool}-only recall ($n=\CctGoldN$), \CctCzeroRec, \CctConeRec\ and
\CctTRec\,\% (seed-level T$-$C0: \CctTCzeroRecSeeds\,pp; T$-$C1:
\CctTConeRecSeeds\,pp), and in the predicted \textsc{call\_tool} rate,
\CctCzeroPredRate, \CctConePredRate\ and \CctTPredRate\,\% against a gold rate of
\CctGoldRate\,\%. C1, whose training labels matched T's action histogram, predicted
\textsc{call\_tool} no more often than C0, so we did not detect label-histogram
absorption as the mechanism for T's higher predicted rate. Macro-F1 over the seven
actions was \CctCzeroMacroF, \CctConeMacroF\ and \CctTMacroF\,\%; T exceeded C0 in
every seed (\CctTCzeroMacroFSeeds\,pp) but C1 in only two of three
(\CctTConeMacroFSeeds\,pp), consistent with the indeterminate $T-C1$ contrast of
Table~\ref{tab:c1}. Because T's recall and predicted rate rose together at roughly
constant precision, these data cannot separate better boundary detection from a
shift toward predicting \textsc{call\_tool} more often; the analysis rests on three
seeds and establishes neither reading.

\begin{table}[t]
\centering
\small
\begin{tabular}{@{}lrrrrr@{}}
\toprule
& \multicolumn{4}{c}{\textsc{call\_tool}} & \\
\cmidrule(lr){2-5}
Arm & P & R & F1 & Pred.\ rate & Macro-F1 \\
\midrule
Gold (n=300) & & & & 29.7 & \\
Random, budget-matched (C0) & 91.9 & 59.2 & 72.0 & 19.1 & 72.8 \\
Random, label-matched (C1) & 91.2 & 56.2 & 69.2 & 18.3 & 72.9 \\
Boundary-targeted (T) & 89.9 & 74.9 & 81.3 & 24.8 & 76.1 \\
\addlinespace
\quad $T-C0$, mean (pp) & ${-}2.0$ & ${+}15.7$ & ${+}9.3$ & ${+}5.7$ & ${+}3.3$ \\
\quad\quad seed signs & +/-/+ & +/+/+ & +/+/+ & +/+/+ & +/+/+ \\
\quad $T-C1$, mean (pp) & ${-}1.3$ & ${+}18.7$ & ${+}12.1$ & ${+}6.4$ & ${+}3.2$ \\
\quad\quad seed signs & -/-/+ & +/+/- & +/+/+ & +/+/- & +/+/- \\
\bottomrule
\end{tabular}

\caption{Per-class view of the C0, C1 and T arms of Table~\ref{tab:c1}
(post-hoc, recomputed from stored pointer-head predictions; \CctNSeeds\ seeds per
arm, \CctNItems\ items, \CctGoldN\ gold \textsc{call\_tool}). Precision, recall,
F1 and predicted rate are for \textsc{call\_tool} only; macro-F1 averages the seven
actions. Seed-mean rows are in \%; contrast rows give the seed-mean difference in
pp and the sign of each seed-paired difference.}
\label{tab:c0c1t}
\end{table}

\begin{table}[t]
\centering
\small
\begin{tabular}{lrrrrrrr}
\toprule
& & \multicolumn{2}{c}{C0} & \multicolumn{2}{c}{C1} & \multicolumn{2}{c}{T} \\
\cmidrule(lr){3-4}\cmidrule(lr){5-6}\cmidrule(lr){7-8}
Gold action & $n$ & P & R & P & R & P & R \\
\midrule
ANSWER & 52 & 91.9 & 94.2 & 93.3 & 84.0 & 91.2 & 98.7 \\
ASK\_CLARIFICATION & 16 & 65.3 & 100.0 & 62.4 & 100.0 & 66.5 & 93.8 \\
ASK\_CONFIRMATION & 24 & 66.2 & 94.4 & 67.5 & 94.4 & 78.4 & 91.7 \\
CALL\_MULTIPLE\_TOOLS & 29 & 40.3 & 63.2 & 47.4 & 59.8 & 58.0 & 55.2 \\
CALL\_TOOL & 89 & 91.9 & 59.2 & 91.2 & 56.2 & 89.9 & 74.9 \\
ESCALATE & 39 & 98.5 & 41.0 & 91.8 & 55.6 & 96.8 & 41.0 \\
REFUSE & 51 & 71.6 & 95.4 & 64.0 & 96.1 & 71.2 & 99.3 \\
\addlinespace
Macro-F1 & & \multicolumn{2}{c}{72.8} & \multicolumn{2}{c}{72.9} & \multicolumn{2}{c}{76.1} \\
\bottomrule
\end{tabular}

\caption{Per-action precision (P) and recall (R) for all seven actions, C0, C1 and
T, in \%, each the mean of the \CctNSeeds\ per-seed values (post-hoc, recomputed
from stored pointer-head predictions on the \CctNItems\ sealed items; $n$ is the
gold count). Precision is taken as 0 for a seed that never predicts the action.
Macro-F1 averages the seven per-action F1 scores. The \textsc{call\_tool} row is
the one summarised in Table~\ref{tab:c0c1t}.}
\label{tab:c0c1tperclass}
\end{table}

\paragraph{A matched-reasoning format experiment we could not factor.}
Section~\ref{sec:twobytwo} leaves a confound we wanted to remove: in the
reasoning-on row the two output interfaces do not share a reasoning prompt. The
obvious fix is to generate the reasoning once, freeze it, and branch only the
output interface. We tried that, in three wirings, and it does not work --- for a
reason worth reporting.

The output instruction changes how long the model reasons. Asked for a single
action name, it reasons for a median of \PfourShortTok\ tokens; given a neutral
prompt with no output instruction, \PfourLongTok, a factor of \PfourTokRatio. The
constrained single-token readout then fails, because after a \PfourLongTok-token
trace the seven action codes are no longer in the top twenty logits at the
decision point --- \PfourBnone\ of \BlindN\ items returned no code at all. Each
wiring works for exactly one interface and collapses for the other: with the
instruction present the single token reaches \PfourAoneTok\,\% and generated JSON
\PfourAjson\,\%; with it absent, JSON reaches \PfourBjson\,\% and the single token
\PfourBoneTok\,\%.

So we could not run the experiment. Changing only the requested output contract
changed the reasoning trajectory itself, and we report that as a failure to
construct the control rather than as a finding about reasoning models. The
mechanism is mundane --- an autoregressive model conditioned on a different
instruction generates a different trace, which is ordinary prompt sensitivity, and
a more patient prompt-engineering effort might yet find a wiring in which the
trace length matches across branches. What it cost us is concrete: within this
harness the reasoning state and the output interface could not be varied
independently, so the format effect in the reasoning-on row of
Section~\ref{sec:twobytwo} keeps the prompt confound we declared there. Nor are
these numbers a performance comparison --- two of the three wirings collapsed
outright, and neither \PfourAjson\,\% nor \PfourBoneTok\,\% can be read as what
that interface is worth. A collapse gate refused all three wirings, which is the
only reason they are in this paragraph rather than in
Section~\ref{sec:twobytwo} as a format effect.

\paragraph{A pre-registered matched-state diagnostic, also void.} In response to
review we ran a second attempt, pre-registered before submission to the GPU. It
froze one reasoning trace per item (the deployed JSON arm's own trace) and read
the seven actions from that fixed state in two ways: the single-token code from
the full-vocabulary logits, and the JSON action strings from teacher-forced
log-probabilities, with no top-$k$ truncation, so the readout failure above could
not recur. It is void. Its positive control failed: when we planted a wrong
concluding sentence at the end of the trace, the readouts followed the planted
action in only \MsWrongTok\,\% (single token) and \MsWrongJson\,\% (JSON) of items,
against a registered threshold of \MsGatePos\,\%; the model overrode the planted
conclusion and re-derived its own action. A readout that does not track the state
it is given cannot attribute a difference to the interface, so we report the
failing gate (Appendix~\ref{app:matchedstate}) and no format effect. The
reasoning-on format effect remains unresolved.

\section{What we claim}
\label{sec:claims}

We state the five claims this paper supports, with the qualifier each one needs,
so that a reader can check them one at a time rather than inferring them from the
sections.

\begin{description}
\item[C1.] On familiar policies, on one 27B Korean enterprise model, we did not
detect an accuracy difference between the cheap single-token operational path and
the deployed reasoning-plus-JSON path (\RzeroConFull\,\% against
\RzeroJsonFull\,\% on the full split), and the cheap path is an order of magnitude
steadier across restarts (\FlipConMean\,\% of decisions change against
\FlipJsonMean\,\%). \emph{Qualifier}: this is a failure to detect a difference, not
an equivalence result, and we ran no paired equivalence test; the comparison does
not isolate the output format from reasoning, because the two paths differ in both;
and the two splits disagree in sign on accuracy (Section~\ref{sec:indist}).

\item[C2.] On unseen policy boundaries, reasoning is decisive. Turning reasoning
on moves exact action from \XOffTok\,\% to \XOnTok\,\%. In the reasoning-off
condition the estimated exact-action difference between the two output formats was
$\XFormatOffDiff\,$pp $[\XFormatOffLo, {+}\XFormatOffHi]$; this is a failure to
detect an advantage for either interface, not an equivalence result
(Section~\ref{sec:twobytwo}). \emph{Qualifier}: the reasoning-on row's two arms do
not share a reasoning prompt, and the post-hoc control of
Section~\ref{sec:posthoc} did not remove that confound, so the reasoning-on format
effect is unresolved. The one format effect this benchmark resolved, on
\textsc{call\_tool}-only recall ($n=\NUnitsTool$), did not reproduce on an open re-rendering of the same
scenarios (Section~\ref{sec:openrerender}).

\item[C3.] Boundary-targeted supervision improves decisions on the boundaries it covers
beyond a token-matched random control --- ${+}\BndDAnsTool\,$pp and
${+}\BndDToolConf\,$pp on the two covered boundaries with headroom --- and we
detect no transfer to a boundary we deliberately withheld
($[{\BndLoRefEsc}, {+}\BndHiRefEsc]\,$pp, an interval that spans zero). \emph{Qualifier}: the treatment arm
also received \textsc{call\_tool} at \PriorToolRatio$\times$ the control's rate,
and a post-hoc label-matched control (Section~\ref{sec:posthoc}) argues against
the class prior as a \emph{sufficient} explanation --- but with one seed flipping
sign it neither conclusively rejects the class prior nor establishes the
construction effect. The size of the one endpoint that
cleared the bar, any-tool recall, also ranges over a factor of \RecallDefSpread\
depending on how the metric is named (Sections~\ref{sec:r3}, \ref{sec:sensitivity} and~\ref{sec:posthoc}).

\item[C4.] Boundary supervision and test-time reasoning are \textbf{partial
substitutes} rather than independent sources of accuracy (ledger verdict:
\RfourVerdict). Measured as a factorial on one harness, reasoning is worth ${+}\EoneDiff\,$pp on the
targeted arm against ${+}\EtwoDiff\,$pp on the random arm; the interaction is
${-}\EfourDiff\,$pp $[\EfourLo, \EfourHi]$ with the same sign in every seed pair
(Section~\ref{sec:ladder}). \emph{Qualifier}: this describes the joint behaviour
of the two budgets in this operating range, not a mechanism --- a common ceiling
remains a partial alternative explanation, accounting for up to \RfourGapOff\ of
the \EfourDiff\,pp under a simple compression model. It is also the one place we
can speak about within-checkpoint reasoning value, and it does not carry across
harnesses --- and on the open re-rendering the interaction keeps its sign but is
not stable (Section~\ref{sec:openrerender}). The pointer-head numbers of
Table~\ref{tab:r3} and of the trained
rows of Table~\ref{tab:ladder} (Sections~\ref{sec:r3} and~\ref{sec:ladder}) come
from a different decision mechanism, not a comparable cell.

\item[C5.] With Korean supervision only, we detect no degradation on English
renderings of the same latent policy scenarios: with reasoning on, English is
${-}\REnXtwo\,$pp $[\REnXtwoLo, {+}\REnXtwoHi]$ from Korean, the same order as
re-rendering in Korean (${-}\REnXthree\,$pp) (Section~\ref{sec:english}).
\emph{Qualifier}: the reasoning-off gain from targeted supervision has similar point
estimates in the two languages (${+}\REnTCKoOff$ and ${+}\REnTCEnOff\,$pp) but is not
stable in either under item-clustered intervals, so we do not claim that it
transfers. No English degradation was \emph{detected}; this is not a claim of
equivalence, it is
measured on a common subset that is easier than the full benchmark, and it does not
cover English-authored enterprise policies or locate where in the lineage the
ability survived.
\end{description}

Everything else in this paper is a measurement we report, a diagnostic we ran, or
a limitation we found. It is not a claim.

\section{Conclusion}

An enterprise agent has to pick an action before it says anything, and we asked what
that pick needs. On one 27B Korean enterprise model and one constructed policy
family, the answer, with the scope and qualifier of each claim, is stated in
Section~\ref{sec:claims}; we do not restate it here.

Two things would most change this picture. The first is evaluation on real
enterprise policies rather than constructed ones, which is the validation this
paper most needs and does not provide. The second is a comparison of output
interfaces under matched reasoning, which our harness could not construct; until it
exists, the effect of the output format with reasoning on remains unresolved.

We deliberately do not propose an adaptive controller here. Our own exploration of
one used the sealed benchmark, which converts it into a development set; the
signals that look most promising on it --- disagreement among cheap non-reasoning
comparators discriminates whether reasoning will help with an AUC of
\UaucDisagree, against \UaucConf\ for model confidence --- are reported in the
pre-registration of a separate experiment rather than claimed here. That
experiment requires a second sealed benchmark, disjoint in template and
vocabulary, and a decision rule fixed before it is opened.

\section*{Artifacts}

The original rendering of the sealed benchmark --- the text two commercial models wrote, on
which every primary analysis rests --- is not redistributed, because we do not have sufficiently clear redistribution rights to
release that generated text under the benchmark's public licence. Its content hash is released,
so its identity as the confirmatory set remains verifiable. To keep the denominators apart: the
\emph{original sealed rendering} (text private) has \BlindN\ items in the main analyses and
\REnN\ in the common subset of the cross-lingual audit; the \emph{open re-rendering} (text
public) has \OpenN\ common items.
What is released is the open re-rendering of Section~\ref{sec:openrerender}: its \OpenN\ common
items in all five audit configurations (Korean, English, the two mixed frame/document variants
and the second Korean rendering), the latent scenario definitions, the
executable gold labels and per-item proofs for all \BlindN\ scenarios, the seals of both
renderings (the original by hash only), the generator, a standalone twelve-check verifier that
depends on none of the code that produced it, and the aggregate measurement ledgers behind
every number in this paper:

\begin{center}
\url{https://huggingface.co/datasets/ThakiCloud/EnterpriseOps-Blind-A}
\end{center}

The same repository carries the English audit of Section~\ref{sec:english}: the
\REnN\ common item identifiers, the seals of the original renderings it used, the
pre-registration and its addendum, the ledger behind Table~\ref{tab:english}, and the
open renderings of every audit configuration (English, Korean frame with English
documents, English frame with Korean documents, and the second Korean rendering). The English items are
independent renderings of the latent scenarios, not translations, and their gold
labels and proofs are those of the Korean set. Having been used for the claims in
this paper, they are no more a blind test set than the Korean items; future
cross-lingual claims need a freshly sealed benchmark.

The base enterprise decision model used for every arm
(\texttt{Qwen3.8-27B-Human-KO-Enterprise-v0.1}) and the boundary-targeted checkpoint of
Section~\ref{sec:r3} are both on the same organisation page:

\begin{center}
\url{https://huggingface.co/ThakiCloud/Qwen3.8-27B-Human-KO-Enterprise-Boundary}
\end{center}

The boundary checkpoint ships as merged full weights with its pointer head rather than as an
adapter, and it keeps the base architecture intact: the merge is applied at the tensor-file
level to the 496 targeted text-tower tensors, leaving the vision tower and the rest
byte-identical. The released boundary checkpoint is
the seed selected on the \emph{in-distribution} evaluation set ($n=5{,}608$); it is not
selected on the sealed benchmark, although it happens also to be the best of the three seeds
there, and we state that so the selection rule is checkable rather than merely asserted.

The training corpus is \emph{not} released. It derives from internal and licensed sources we
are not in a position to redistribute. The released renderings were produced with
open-weight models (Qwen3.6-27B and Gemma-4-31B); commercial models acted only as gate judges
and wrote none of the released text.

Releasing the benchmark makes this paper checkable and makes the benchmark contaminable. We
think that trade is correct here because we have already used it to develop an adaptive
method, which disqualifies it as a test set for our own future claims regardless of whether we
publish it. The decision rule and the second sealed benchmark that adaptive work will be
judged on are pre-registered separately, before that benchmark exists.

\bibliographystyle{plainnat}
\bibliography{refs}

\appendix

\section{Measurement index}
\label{app:index}

Every result table and figure in the main text, with the checkpoint and interface it
measures, its sample size and denominator, the number of runs or seeds, how its
uncertainty was computed, and the harness that produced it. The table is generated
from \texttt{drafts/measurement\_index.json}, which reads its values from the
measurement ledgers. \CiMethodCaption\ Single-run tables (Table~\ref{tab:twobytwo})
use a paired item-resampling bootstrap over the one run, as listed.

{\scriptsize
\begin{longtable}{@{}>{\raggedright\arraybackslash}p{0.13\linewidth}>{\raggedright\arraybackslash}p{0.16\linewidth}>{\raggedright\arraybackslash}p{0.08\linewidth}>{\raggedright\arraybackslash}p{0.12\linewidth}>{\raggedright\arraybackslash}p{0.15\linewidth}>{\raggedright\arraybackslash}p{0.14\linewidth}>{\raggedright\arraybackslash}p{0.14\linewidth}@{}}
\toprule
Result & Checkpoint / interface & $n$ & Runs / seeds & Denominator & Uncertainty & Harness \\
\midrule
\endfirsthead
\toprule
Result & Checkpoint / interface & $n$ & Runs / seeds & Denominator & Uncertainty & Harness \\
\midrule
\endhead
\bottomrule
\endfoot
Tab.~\ref{tab:indist}\newline Reasoning+JSON vs no-reasoning single token, in-distribution & Qwen3.8-27B-Human-KO-Enterprise-v0.1 (no adapter)\newline \emph{JSON action object (deployed path) vs constrained single token (letter code)} & clean 2414; full 2800 & json restarts 5; json pairs 10; constrained restarts 4; constrained pairs 6; accuracy reps used 2,3,4,5\newline no training seeds; restarts of the same serving checkpoint (temperature 0) & accuracy: strict clean split 2414 / full 2800 items; flip: same two splits, items whose action differs between two restarts & none pooled; mean/min/max over restart pairs (pairs share restarts, not independent) & Section 3 harness (pace/r0, vLLM v0.28.0, H200 x1); NOT the Section 6 pointer-head harness \\
\addlinespace
Tab.~\ref{tab:twobytwo}, Fig.~\ref{fig:twobytwo}, Fig.~\ref{fig:twobytwocf}\newline Reasoning x output format 2x2 on sealed Blind-A & v0.1 base (no adapter)\newline \emph{\{reasoning on/off\} x \{JSON object, single token\}} & items 300; cf pairs 30; invariance pairs 15 & 1 run per cell\newline single run per cell (no seeds); temperature 0 & exact action: all 300 items; both-correct: 30 CF pairs (pair is the unit); call\_tool\_recall (secondary): 89 CALL\_TOOL-only items & paired item bootstrap, n\_boot=2000, fixed seed 20261001; pair-level resampling for pair metrics & Section 5 serving harness pace/race/race\_2x2.py (vLLM v0.28.0, B200 x1, concurrency 16) \\
\addlinespace
Tab.~\ref{tab:r3}, Fig.~\ref{fig:boundaries}\newline Boundary-targeted (T) vs token-budget-matched random (C) supervision & v0.1 + LoRA r=16 + pointer head, 3 seeds per arm\newline \emph{trained pointer head (argmax over 7 options), no generation, no reasoning} & items 300; cf pairs 30; any tool required 118; boundary pooled 128; invariance pairs 15 & 3 seeds x 2 arms\newline 1,2,3 (T\_i vs C\_i seed-paired) & Recall endpoint here = ANY-tool-required items n=118 (CALL\_TOOL + CALL\_MULTIPLE\_TOOLS); the 2x2 ledger CALL\_TOOL-only n=89 & item-clustered bootstrap (item\_clustered\_bootstrap\_v1): units with replacement; all seeds of a unit stay together; B=10000, seed 12345; stable = unchanged registered definition: per-seed signs agree (len(\{d>0\})==1) AND CI excludes 0 & Section 6 pointer-head harness (kev eval\_kev.py, LocalPredictor + head.pt); trained T-s1/C-s1/T-s3/C-s3 on H200 NVL, T-s2/C-s2 on B200 \\
\addlinespace
Tab.~\ref{tab:ladder}, Fig.~\ref{fig:ladder}\newline Configuration ladder (competing options) & mixed: v0.1 base (rows 1,5), R1b pointer head (row 2), R3 C and T adapters (rows 3,4)\newline \emph{row1 vLLM constrained token; row2-4 pointer head; row5 vLLM reasoning-on single token} & items 300 & row1 1; row2 3; row3-4 3; row5 1\newline rows 2-4 seed means (3 each); rows 1,5 single run & exact action over 300 & no intervals in the table; bars = seed min/max in figure & MIXED: rows 1,5 = Section 5 vLLM probe (B200); rows 2,3,4 = Section 6 pointer-head harness \\
\addlinespace
Tab.~\ref{tab:r4cells}, Fig.~\ref{fig:r4}\newline Training x reasoning, 4 cells, one harness & v0.1 + LoRA r=16 (R3 adapters, pointer head unused), 3 seeds per arm, key-fixed for vLLM\newline \emph{vLLM LoRA serving; no-think = enable\_thinking=false 1-token; think = on\_1tok two-stage} & items 300 & 12 cells (6 adapters x 2 reasoning states)\newline [1, 2, 3] & exact action over 300; recalls secondary (CALL\_TOOL-only 89 / any-tool 118) & item-clustered bootstrap (item\_clustered\_bootstrap\_v1): units with replacement; all seeds of a unit stay together; B=10000, seed 12345; stable = unchanged registered definition: per-seed signs agree (len(\{d>0\})==1) AND CI excludes 0 & Section 7 serving harness (one vLLM v0.28.0 server, --enable-lora, H200 x1 per r6en ledger reproduction\_check.note; spec pace/r4/job/pace-r4-serve-h200.json) \\
\addlinespace
Tab.~\ref{tab:sensitivity}\newline Denominator and shortcut sensitivity (recompute only) & as in Tables 2,3\newline \emph{as in Tables 2,3} & all 300; shortcut clean 240; call tool only 89; any tool 118 & reuses Table 2/3 predictions\newline as source & CALL\_TOOL-only n=89 vs any-tool n=118; clean layers = 5 of 7 (dialect\_noisy, missing\_information dropped) & point estimates only (P2 ledger has no CI) & mixed: 2x2 effects = Section 5 serving; T-C = Section 6 pointer head \\
\addlinespace
Tab.~\ref{tab:english}\newline Zero-shot English transfer (frame x document language) & v0.1 + R3-T LoRA (3 seeds) for Korean/English cells; Table also T vs C contrasts\newline \emph{vLLM LoRA, 1-token, reasoning on/off} & items 286; note common intersection of 4 sets; ko-all-rows n=300 reported separately (selection\_effect) & 5 sets x 2 reasoning x 3 seeds\newline [1, 2, 3] & 286 items common to ko/en/endoc/enframe/ko2 & item-clustered bootstrap (item\_clustered\_bootstrap\_v1): units with replacement; all seeds of a unit stay together; B=10000, seed 12345; stable = unchanged registered definition: per-seed signs agree (len(\{d>0\})==1) AND CI excludes 0 & Section 7 serving harness (R6en, B200 x1, spec pace/r6en/job/pace-r6en-serve-b200.json) \\
\addlinespace
Tab.~\ref{tab:replication}\newline Open re-rendering replication (post-seal robustness) & same as Tables 2-5,9\newline \emph{same} & open common 248; original full 300 & re-scoring of existing checkpoints on open rendering\newline same as source tables & 248 common intersection items; original restricted rows exist for subset control & item-clustered bootstrap (item\_clustered\_bootstrap\_v1): units with replacement; all seeds of a unit stay together; B=10000, seed 12345; stable = unchanged registered definition: per-seed signs agree (len(\{d>0\})==1) AND CI excludes 0 & Section 5/6/7 harnesses re-run on rendering 'inhouse' (B200 evalr-1..4, ptr-a/b) \\
\addlinespace
Tab.~\ref{tab:c1}\newline Label-matched random control (C1) & v0.1 + LoRA r=16 + pointer head; C1 = 3 new trained seeds (H200 NVL)\newline \emph{pointer head (same as Table 3)} & items 300; boundary layers n 120; boundary layers 3 & 3 seeds x 3 arms (T, C0 reused from R3, C1 new)\newline 1,2,3 & boundary layers = 3 layers (call\_tool\_boundary, confirm\_refuse\_boundary, missing\_information) n=120 -- NOT the 128 pooled boundary groups of Table 3; CALL\_TOOL-only recall n=89 & item-clustered bootstrap (item\_clustered\_bootstrap\_v1): units with replacement; all seeds of a unit stay together; B=10000, seed 12345; stable = unchanged registered definition: per-seed signs agree (len(\{d>0\})==1) AND CI excludes 0 & Section 6 pointer-head harness \\
\addlinespace
\end{longtable}
}

\section{Reproducibility details}
\label{app:repro}

\paragraph{Models and data.} Every arm starts from
\texttt{Qwen3.8-27B-Human-KO-Enterprise-v0.1} in bf16. The sealed benchmark is
identified by its content hash (\texttt{\BlindSha}\ldots); the Korean configuration
of the open re-rendering by \texttt{\OpenKoShaPrefix}\ldots\ over its
\OpenKoSealN\ items.

\begin{figure}[ht]
\centering
\includegraphics[width=\linewidth]{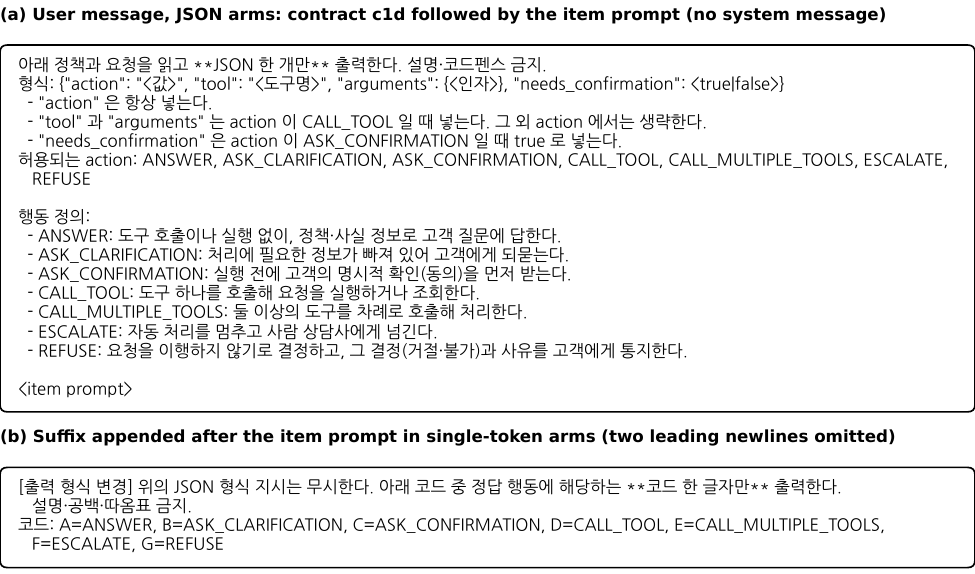}
\caption{Verbatim Korean prompt contract (a) and single-token suffix (b), rendered
from the code that built the requests. Lines too long for the page are wrapped
with an indent; no other change is made.}
\label{fig:prompts}
\end{figure}

\paragraph{Prompts.} No system message is sent on any path; the whole prompt is
one user message. For the JSON arms it is the contract of
Figure~\ref{fig:prompts}(a) followed by the item text (policy, tools, session
notes and request). The single-token arms append the suffix of
Figure~\ref{fig:prompts}(b). Both are reproduced verbatim from the source that
built the requests. The English-frame cells of Table~\ref{tab:english} use the
English contract and suffix below, which translate the Korean ones line for line
and leave the action names untranslated.

\begin{Verbatim}[fontsize=\footnotesize,breaklines,breakanywhere,frame=single]
Read the policy and the request below and output **exactly one JSON object**. No explanation, no code fences.
Format: {"action": "<value>", "tool": "<tool name>", "arguments": {<arguments>}, "needs_confirmation": <true|false>}
  - "action" is always included.
  - "tool" and "arguments" are included when action is CALL_TOOL. They are omitted for every other action.
  - "needs_confirmation" is included as true when action is ASK_CONFIRMATION.
Allowed actions: ANSWER, ASK_CLARIFICATION, ASK_CONFIRMATION, CALL_TOOL, CALL_MULTIPLE_TOOLS, ESCALATE, REFUSE

Action definitions:
  - ANSWER: answer the customer's question from policy and known facts, with no tool call and no execution.
  - ASK_CLARIFICATION: information needed to proceed is missing, so ask the customer back.
  - ASK_CONFIRMATION: obtain the customer's explicit confirmation (agreement) before executing.
  - CALL_TOOL: call one tool to execute or look up the request.
  - CALL_MULTIPLE_TOOLS: call two or more tools in turn to handle it.
  - ESCALATE: stop automatic handling and hand it to a human agent.
  - REFUSE: decide not to carry out the request, and tell the customer that decision (refusal, not possible) and the reason.
<item prompt>
\end{Verbatim}
\noindent Single-token suffix (appended after the item prompt in single-token arms; leading blank lines shown as-is):
\begin{Verbatim}[fontsize=\footnotesize,breaklines,breakanywhere,frame=single]
[Output format change] Ignore the JSON format instruction above. Output **only the single letter code** among the codes below that corresponds to the correct action. No explanation, no whitespace, no quotes.
Codes: A=ANSWER, B=ASK_CLARIFICATION, C=ASK_CONFIRMATION, D=CALL_TOOL, E=CALL_MULTIPLE_TOOLS, F=ESCALATE, G=REFUSE
\end{Verbatim}

Requests set only \texttt{temperature 0}; \texttt{top\_p}, \texttt{seed} and
\texttt{stop} are unset, and \texttt{max\_tokens} is unset for generation and one
for single-token reads. Reasoning is switched per request through
\texttt{chat\_template\_kwargs.enable\_thinking}. A single-token decision is the
argmax over the action codes present among the top-20 logprobs at the first
position. With reasoning on, the single-token arm runs in two stages: a chat request
produces the reasoning, and a completion request then reads one token after the
reasoning and a closing think tag; an over-long trace is cut token-wise to fit the
context. The pointer head generates nothing: a trained head scores the seven
actions from the encoded state and the argmax is taken.

\paragraph{Action codes.} Action names are not single tokens, and two of them share
their first token, so the constrained arms use one-letter codes
(Table~\ref{tab:app-tokens}). The decision is read by exact match of the logprob
token string to the code.

\begin{table}[ht]
\centering
\small
\begin{tabular}{lcrr}
\toprule
Action & Code & Token id & Tokens in action name \\
\midrule
ANSWER & \texttt{A} & 32 & 2 \\
ASK\_CLARIFICATION & \texttt{B} & 33 & 4 \\
ASK\_CONFIRMATION & \texttt{C} & 34 & 3 \\
CALL\_TOOL & \texttt{D} & 35 & 2 \\
CALL\_MULTIPLE\_TOOLS & \texttt{E} & 36 & 5 \\
ESCALATE & \texttt{F} & 37 & 3 \\
REFUSE & \texttt{G} & 38 & 2 \\
\bottomrule
\end{tabular}

\caption{Action, letter code and token id in the Qwen3.8 tokenizer, with the
number of tokens in each action name. The same mapping was used on every restart.}
\label{tab:app-tokens}
\end{table}

\paragraph{Training.} The boundary-targeted, budget-matched random and
label-matched arms share one set of training arguments (Table~\ref{tab:app-train});
only the data differ.

\begin{table}[ht]
\centering
\small
\begin{tabular}{@{}lp{0.62\linewidth}@{}}
\toprule
Item & Value \\
\midrule
Base checkpoint & Qwen3.8-27B, revision 1d4bf0f2 \\
Adapter & LoRA rank 16, alpha 32, dropout 0.05, bias none \\
LoRA targets & down\_proj, gate\_proj, in\_proj\_a, in\_proj\_b, in\_proj\_qkv, in\_proj\_z, k\_proj, o\_proj, out\_proj, q\_proj, up\_proj, v\_proj \\
Pointer head & hidden dim 256, trained from scratch \\
Trainable parameters & 119,349,760 \\
Learning rate & LoRA 5e-5, head 1e-3 \\
Epochs / batch & 1 epoch; batch 4 $\times$ accumulation 4 = 16 records per step \\
Optimizer steps & targeted 412, random 417, label-matched 412 \\
Training records & targeted 6,588, random 6,666, label-matched 6,588 (base 3,988) \\
Added tokens & targeted 1,141,753, random 1,138,741 (-0.26\%), label-matched 1,124,979 (-1.47\%) \\
Max state length & 1024 tokens \\
Precision & bf16 compute, bf16 frozen weights, gradient checkpointing on \\
Peak device memory & 56.7--56.9 GB over the 9 runs \\
Seeds & 1, 2, 3 per arm, seed-paired across arms \\
\bottomrule
\end{tabular}

\caption{Training configuration of the three supervision arms (boundary-targeted,
budget-matched random, and the post-hoc label-matched arm of
Section~\ref{sec:posthoc}), read from each arm's job specification, its token-budget
record, the adapter configuration and the per-run training metrics. The build
asserts that all nine job specifications share these training arguments and base
revision and that the adapters match.}
\label{tab:app-train}
\end{table}

\paragraph{Serving.} All vLLM measurements use the engine configuration of
Table~\ref{tab:app-serve}, read back from the running server. The LoRA cells of
Tables~\ref{tab:r4cells} and~\ref{tab:english} add LoRA serving with all six
adapters loaded on one server. The adapter keys had to be rewritten for vLLM: in a
first attempt no module matched and the output was identical to the base model, so
that attempt measured the base checkpoint and was discarded.

\begin{table}[ht]
\centering
\small
\begin{tabular}{@{}ll@{}}
\toprule
Flag & Value \\
\midrule
\texttt{Engine image} & \texttt{vllm/vllm-openai:v0.28.0} \\
\texttt{dtype} & \texttt{bfloat16} \\
\texttt{max-model-len} & \texttt{8192} \\
\texttt{max-num-seqs} & \texttt{256} \\
\texttt{gpu-memory-utilization} & \texttt{0.9} \\
\texttt{kv-cache-dtype} & \texttt{fp8} \\
\texttt{enable-prefix-caching} & \texttt{true} \\
\texttt{reasoning-parser} & \texttt{qwen3} \\
\texttt{tensor-parallel-size} & \texttt{1} \\
\texttt{TORCH\_COMPILE\_DISABLE} & \texttt{0} \\
\bottomrule
\end{tabular}

\caption{vLLM engine configuration, identical across the in-distribution and
sealed-benchmark harnesses (checked against both readbacks).}
\label{tab:app-serve}
\end{table}

\paragraph{Hardware.} The in-distribution measurements of
Section~\ref{sec:indist} ran on one H200; the two-by-two and the race of
Section~\ref{sec:twobytwo}, the cross-lingual audit and the open re-rendering on
one B200; the factorial of Section~\ref{sec:ladder} on one H200; the label-matched
control on one H200 NVL. The supervision arms of Section~\ref{sec:r3} trained on
H200 NVL for seeds 1 and 3 and on B200 for seed 2, always with both arms of a seed
pair on the same hardware. Absolute latencies are therefore not comparable across
sections.

\paragraph{Latency.} Table~\ref{tab:app-latency} gives absolute latencies and
generated tokens. The in-distribution rows are medians over restarts of the
cold-prefix cells; the sealed-benchmark rows are wall-clock per request at
concurrency 16 and include queueing. The percentage cuts quoted in
Section~\ref{sec:indist} are computed by the measurement script from these cells.

\begin{table}[ht]
\centering
\small
\begin{tabular}{@{}llrrr@{}}
\toprule
Setting & Path & p50 (s) & p95 (s) & Median gen.\ tokens \\
\midrule
In-distribution, $c=1$ & Reasoning + JSON (cold prefix) & 8.89 & 27.43 & 538 \\
 & No reasoning, single token (cold prefix) & 0.085 & 0.168 & 1 \\
In-distribution, $c=8$ & Reasoning + JSON (cold prefix) & 10.68 & 27.71 & 568 \\
 & No reasoning, single token (cold prefix) & 0.107 & 0.173 & 1 \\
\addlinespace
Sealed benchmark, $c=16$ & Reasoning on, JSON & 12.55 & 68.27 & 972 \\
 & Reasoning on, single token (both stages) & 4.19 & 41.25 & 255 \\
 & Reasoning off, JSON & 0.62 & 2.11 & 14 \\
 & Reasoning off, single token & 0.61 & 0.74 & 1 \\
\bottomrule
\end{tabular}

\caption{Absolute latency and median generated tokens (reasoning included). In
distribution: \AbsLatN-item sample, \AbsLatReps\ restarts. Sealed benchmark: all
\BlindN\ items, one run per path; the reasoning-on JSON latency comes from a rerun,
while its accuracy in Table~\ref{tab:twobytwo} comes from the original run.}
\label{tab:app-latency}
\end{table}

\paragraph{Not recorded.} We did not record: the rendered generation-prompt tail
with reasoning on; a hash of the chat template; the latency of the pointer-head
path; the GPU type of two of the pointer-baseline seeds; the node that ran the
post-hoc matched-format experiment; and scores for space-prefixed letter codes.

\section{Measurement caveats and audit trail}
\label{app:caveats}

These caveats do not change the claims of Section~\ref{sec:claims}, but each
matters for checking a particular number.

The English audit of Section~\ref{sec:english} inherits the Korean-first
construction of the benchmark (Section~\ref{sec:bench}). The latent fact
specifications from which both languages were rendered were authored in
Korean. Because no English enterprise supervision was used, this gives the model no
direct English label-memorisation path, but it may under-represent English-specific
policy formulations and discourse conventions --- Anglophone compliance terminology,
contract register, jurisdiction-specific wording, culturally specific escalation
language --- none of which this benchmark measures. The audit also stopped where its
pre-registration said it would: because the retention gate passed, we did not
evaluate the base model or the intermediate Korean checkpoint in English, and so we
cannot say at which stage of the lineage English decision ability was preserved. It
shows that the released checkpoint transfers, not why.

The interaction we call partial substitution cannot be separated from a ceiling.
Both reasoning-on cells sit near \RfourCThink\,\%, which is the highest exact-action
figure anywhere in this paper, and a ceiling at that level would compress the two
arms together and produce a negative interaction on its own. Section~\ref{sec:ladder}
bounds how much of the effect that could account for --- at most \RfourGapOff\ of
\EfourDiff\,pp, the share that pure compression can remove --- but bounding is not
excluding. Establishing a ceiling would require an arm that clears it, and we do not
have one. ``Partial substitutes'' is therefore a description of how the two budgets
behave together in this operating range, not an identified mechanism.

Two endpoints in Table~\ref{tab:r3} report differences smaller than the data can
resolve. One counterfactual pair is worth \OneUnitCf\,pp and one invariance pair
\OneUnitInv\,pp, so the ${+}\BndDMiss\,$pp on missing information and the
${+}\BndDInv\,$pp on invariance pairs are each below a single unit. We report them
for completeness and do not interpret them; the ceiling reading we offer for
missing information in Section~\ref{sec:r3} is an interpretation we did not
pre-register and that the resolution does not support on its own.

We report no in-distribution accuracy difference without having run a paired test,
and the two splits disagree in sign: on the full split the single token is ahead
by 0.68\,pp (19 items), on the clean split generated JSON is ahead by 0.11\,pp.
Neither direction is established.

The pre-registered kill switch guarded against a different failure
(constant \textsc{refuse} on the policy-versus-principle slice) and would not
have fired on this one. Record counts and optimiser steps also differ slightly
between arms (\RecordsTreat\ against \RecordsCtrl\ records, \StepsTreat\ against
\StepsCtrl\ steps); only the token budget was matched, to within
\BudgetSkewPct\,\%.

The two paths were also not evaluated over the same restart set. The deployed
path was measured across \RestartsJson\ restarts including the first boot, the
single-token path across \RestartsCon\ later ones, because the first
single-token cycle was pre-empted. A first-boot effect would therefore land on
one arm only.

One of our pre-registered gates could not fire. We asked, before training, that
per-arm seed noise land in a 1.5--3.0\,pp band, as a check that the runs were
neither suspiciously identical nor wildly unstable. Observed noise ranges from
\KfiveSdLo\ to \KfiveSdHi\,pp, so \KfiveInBand\ of \KfiveArms\ endpoints fell in
the band and all of them fell above it. A two-sided band on a quantity we had no
prior for was the wrong shape: the question we actually wanted answered is
whether noise exceeds a ceiling, and a band that every arm overshoots carries no
information. We leave the gate as registered rather than re-fitting it after
seeing the data, and record the correct form here for the next registration.

Two of our own ledgers compute \textsc{call\_tool} recall over different
denominators. The two-by-two ledger counts the \NUnitsTool\ items whose gold
action is exactly \textsc{call\_tool}; the boundary-supervision ledger counts
\NUnitsToolRthree, folding in \textsc{call\_multiple\_tools}. Both are defensible
readings of ``a tool call is required'' and neither is wrong, but they share a
metric name, so the two numbers are not comparable across sections and an
inattentive reader --- or an inattentive author --- will compare them. We caught
this only while making every number in this paper regenerate from its ledger, and
we state it because the endpoint it affects is the one endpoint that passed.

The counterfactual endpoint rests on \BlindCfPairs\ pairs, and its seed-to-seed
spread in the supervision experiment is large. We report it, we do not rest a
claim on it.

The pointer head's single-pair restart measurement is not a stability claim: that
pair was bit-identical, and we could not separate deterministic kernels from
caching with one pair.

Finally, every number here is the version of the system we measured. The
measurement ledgers backing each figure and table are released with the paper so
that a disagreement can be a disagreement about data rather than about recall.

The flip statistics of Section~\ref{sec:indist} use all \RestartsJson\ JSON
restarts, while its accuracy means and latencies use only the later restarts. The
latency sample is \AbsLatN\ items per restart, so its p95 is a noisy estimate. The
headline percentage cuts come from the measurement script's decode-share estimate;
the plain ratio of the cells in Table~\ref{tab:app-latency} gives the same values
after rounding at p50 and differs slightly at p95.

\subsection{Matched-state diagnostic (pre-registered, void)}
\label{app:matchedstate}

\emph{Design.} For each of the \MsN\ sealed items we generated one reasoning trace
under the deployed JSON contract (reasoning on, greedy) and froze its tokens. From
that fixed state we read the seven actions twice, after a follow-up user turn that
differed only in the requested output form: the single-token code by argmax over
the full-vocabulary logits at the next position, and the JSON action by the summed
teacher-forced log-probability of each of the seven \texttt{\{"action": ...\}}
strings. Four states were scored: the item's own trace (real); the trace with an
appended sentence stating the gold action (planted correct) or a randomly chosen
wrong action (planted wrong), as positive controls; and the trace of a different
item (shuffled), as a negative control. The registered primary contrast was the
paired single-token minus JSON accuracy on the real state, with five gates fixed
before the run (Table~\ref{tab:app-matchedstate}); a failure of G1, G2 or G4
voids the run.

\begin{table}[h]
\centering\small
\begin{tabular}{@{}p{0.30\linewidth}p{0.36\linewidth}rc@{}}
\toprule
Gate & Criterion & Value & Pass \\
\midrule
G1 positive control, planted correct & both readouts $\geq 90$\,\% follow the planted action & 93.0 / 97.7 & yes \\
G1 positive control, planted wrong & both readouts $\geq 90$\,\% follow the planted action & 10.3 / 50.0 & \textbf{no} \\
G2 no collapse (real) & both readouts $\geq 40$\,\% & 85.7 / 86.0 & yes \\
G3 trace sensitivity & real ${-}$ shuffled $\geq 3$ pp in either readout & 67.7 / 66.3 pp & yes \\
G4 determinism & re-scored argmax agreement $\geq 95$\,\% & 100.0 & yes \\
G5 truncation & truncated traces $\leq 5$\,\% & 3.3 & yes \\
\bottomrule
\end{tabular}

\caption{Gates of the matched-state diagnostic (generated from the measurement
ledger). Paired values are single token / JSON, in \% of \MsN\ items; for the
planted conditions the value is the share of items whose readout equals the
planted action. G4 re-scores a subset of the real items. G1 fails on the
planted-wrong condition, so the run is void.}
\label{tab:app-matchedstate}
\end{table}

\emph{Outcome.} Void: G1 failed. The registered primary difference was
\MsPrimaryDiff~pp $[\MsPrimaryLo, \MsPrimaryHi]$; under the pre-registered decision
rule it is not interpretable, and we do not cite it as a format effect or as
evidence of equivalence. A readout that does not follow a planted conclusion
cannot show that two interfaces read the same state the same way.

\emph{Post-hoc observation (descriptive only, not a registered endpoint).} Under
the planted-wrong state the two readouts followed the planted action at different
rates, \MsWrongTok\,\% for the single token and \MsWrongJson\,\% for JSON, while
under the real state they were close (\MsRealTok\,\% and \MsRealJson\,\%). This
suggests that the two readouts are not equally sensitive to the frozen state, which
is itself a reason a matched-state comparison of them would be hard to interpret.

\emph{Threats.} Keeping the reasoning of an earlier turn in the history is
unnatural input: the chat template normally removes it, and the positive control
only partly protects against the two readouts being affected differently.
Teacher-forced scoring also removes the parse failures of free generation, so even
a valid run would have compared choices from a fixed state, not deployed
generation.

\section{Changes in v2}
\label{app:changes}

\begin{itemize}
\item Claim wording: the in-distribution comparison and the reasoning-off format
contrast are stated as failures to detect a difference, not as equivalence
(Abstract, Contributions, Sections~\ref{sec:indist} and~\ref{sec:twobytwo}, C1, C2).
Other statements of absence in the body are worded the same way (``we did not
detect''), and the confounded reasoning-on row is described as unresolved rather
than as a null.
\item The reasoning-on output-format effect is marked unresolved: its two arms do
not share a reasoning prompt, the post-hoc control did not resolve it, and a new
pre-registered matched-state diagnostic (Section~\ref{sec:posthoc},
Appendix~\ref{app:matchedstate}) was void because its positive control failed;
its primary difference is reported in the appendix only and is not interpreted.
\item The training-by-reasoning interaction is stated as stable on the sealed
rendering only, in the Contributions as in the Abstract and C4.
\item New worked benchmark example (Figure~\ref{fig:example}), an explicit
statement of what the executor does and does not prove, and a definition of the
recovery and challenge gates where they are first used (Section~\ref{sec:bench}).
\item New per-class analysis of the targeted, random and label-matched arms:
\textsc{call\_tool} precision, recall and predicted rate with seed-level
differences (Table~\ref{tab:c0c1t}), the marginal predicted-action distribution
(Figure~\ref{fig:c0c1t}), the full $7{\times}7$ confusion matrices
(Figure~\ref{fig:c0c1tcm}) and per-action precision and recall for all seven actions
(Table~\ref{tab:c0c1tperclass}). The Limitations section now points to this
post-hoc analysis instead of saying these quantities were not available.
\item Confidence intervals for every seed-averaged paired comparison, including the
pre-registered lower bounds of Table~\ref{tab:r3} and the per-boundary intervals of
Figure~\ref{fig:boundaries}, now use one item-clustered bootstrap with items in a
fixed order; earlier versions treated each item as three independent draws for some
tables, and resampled the boundary-supervision endpoints with different bootstrap
parameters. The same contrast now carries the same interval in every table (the
exact-action $T-C0$ contrast of Tables~\ref{tab:r3}, \ref{tab:replication}
and~\ref{tab:c1}), and the build checks this. The pre-registered verdict of
Section~\ref{sec:r3} is unchanged. Two conclusions are weakened: the reasoning-off
targeted-supervision gain in the cross-lingual audit is no longer stable in Korean
or English, and the training-by-reasoning interaction no longer replicates on the
open re-rendering (it remains stable on the sealed set).
\item New measurement index (Appendix~\ref{app:index}) and reproducibility appendix
(Appendix~\ref{app:repro}) with verbatim prompts, the action--token mapping,
training configuration of all three supervision arms, serving configuration,
hardware and absolute latency; absolute latency is also given in
Section~\ref{sec:indist}.
\item Limitations reduced to five core items; the other caveats moved to
Appendix~\ref{app:caveats}. The Conclusion points to Section~\ref{sec:claims}
instead of restating the claims.
\item Label fixes: every recall over the $n=\NUnitsTool$ items whose gold action is
exactly \textsc{call\_tool} is called \textsc{call\_tool}-only recall, and every
recall over the $n=\NUnitsToolRthree$ tool-requiring items is called any-tool
recall, in the text and in Tables~\ref{tab:c1} and~\ref{tab:replication}; the
pointer-head figure in Section~\ref{sec:twobytwo} is marked as a
\RacePtrNSeeds-seed mean; the captions of Tables~\ref{tab:ladder}
and~\ref{tab:r4cells} name their harnesses; a wrong section reference in C4 and a
count in Section~\ref{sec:openrerender} were corrected.
\end{itemize}

\end{document}